\documentclass[preprint,12pt]{elsarticle}

\usepackage{amsmath,amssymb,amsfonts}
\usepackage{bm}
\usepackage{marvosym}

\usepackage{colortbl}
\usepackage{xcolor}
\usepackage{array}
\usepackage{diagbox}

\usepackage{manyfoot}
\usepackage{booktabs}
\usepackage{longtable}
\usepackage{siunitx}

\usepackage{algorithm}
\usepackage{algorithmicx}
\usepackage{algpseudocode}

\usepackage{listings}
\usepackage{graphicx}

\usepackage{subcaption}
\usepackage{placeins}

\usepackage{tikz}
\usetikzlibrary{arrows.meta}
\tikzset{>=latex} 

\usepackage{geometry}
\usepackage{microtype}

\usepackage{hyperref}
\hypersetup{
    colorlinks=true,
    linkcolor=black,
    filecolor=black,
    urlcolor=black,
    citecolor=black
}
\journal{  }
\begin{document}
\begin{frontmatter}

\title{\textcolor{black}{Architecture--Optimization Co-Design for Physics-Informed Neural Networks via Layer-wise Coordinate Adaptation and Gradient Conflict Resolution}}

\author{Pancheng Niu \textsuperscript{a}}
\author{Jun Guo \textsuperscript{a,b,\Letter}}
\author{Qiaolin He \textsuperscript{c}}
\author{Yongming Chen \textsuperscript{a}}
\author{Yanchao Shi \textsuperscript{d}}

\affiliation{organization={College of Applied Mathematics},
            addressline={Chengdu University of Information Technology},
            city={Chengdu},
            postcode={610225},
            state={Sichuan},
            country={P. R. China}}

\affiliation{organization={Key Laboratory of Mathematical Meteorology},
            addressline={Chengdu University of Information Technology of Sichuan Province},
            city={Chengdu},
            postcode={610225},
            state={Sichuan},
            country={P. R. China}}

\affiliation{organization={College of Mathematics},
            addressline={Sichuan University},
            city={Chengdu},
            postcode={610064},
            state={Sichuan},
            country={P. R. China}}

\affiliation{organization={School of Sciences},
            addressline={Southwest Petroleum University},
            city={Chengdu},
            postcode={610500},
            state={Sichuan},
            country={P. R. China}}       
\begin{abstract}

{\color{black}Physics-informed neural networks (PINNs) can be limited by how physical coordinates are mapped to latent features and by gradient conflicts among heterogeneous physical constraints. We propose Architecture--Conflict-Resolved PINN (ACR-PINN), which integrates Layer-wise Dynamic Adaptation (LDA) with Gradient-Conflict-Resolved PINN (GCR-PINN) to address these coupled sources of training difficulty. LDA performs layer-wise coordinate adaptation by constructing layer-specific coordinate features and fusing two encoding branches through input-conditioned, feature-wise gates. GCR-PINN treats the governing-equation residual, initial-condition, and boundary-condition losses as separate optimization tasks and conditionally projects negatively aligned task gradients before aggregation. ACR-PINN thereby modifies both feature construction and gradient aggregation without changing the underlying physical loss formulation. Across seven benchmark problems, it attains the lowest observed mean across all reported error metrics among the four core architecture--optimizer configurations. Relative to a standard PINN (Std-PINN), ACR-PINN reduces the mean relative $L_2$ error by approximately 56--97\% and consistently improves upon the better-performing single-component configuration. Controlled analyses further clarify the source of these gains. Parameter-matched comparisons and targeted gate interventions indicate that the LDA improvement reflects the learned coordinate-fusion pathway rather than merely a larger number of trainable parameters. Gradient diagnostics further suggest that the resulting representation can improve task-gradient compatibility and reduce the correction required by GCR-PINN, although the magnitude of this effect varies across problems and update rules. Comparisons with representative PINN methods provide additional support for the observed accuracy gains, while robustness tests show that these gains persist under the examined variations in sampling and temporal horizon. Overall, the results indicate that jointly adapting coordinate features and resolving task-gradient conflicts can improve PINN accuracy across the problems considered.}

\end{abstract}

\end{frontmatter}

\section{Introduction}\label{sec1}

Physics-informed neural networks (PINNs) solve partial differential equations (PDEs) by incorporating physical laws into neural network optimization \cite{raissi2019physics, karniadakis2021physics}. The governing equations, together with initial and boundary conditions, are usually enforced through residual terms in the loss function. With automatic differentiation, the required derivatives can be evaluated without constructing conventional numerical meshes or discrete differential operators, allowing PINNs to address both forward and inverse problems in a mesh-free setting \cite{baydin2018automatic}. This formulation has been applied to a range of scientific and engineering problems, including fluid dynamics \cite{jin2021nsfnets, jagtap2022physics}, heat transfer \cite{cai2021physics, xu2023physics}, solid mechanics \cite{haghighat2021physics, hu2024physics}, acoustic and elastic wave propagation \cite{rasht2022physics, alkhadhr2023wave}, and periodic water-wave modeling \cite{chen2024wavenets}.

Despite their broad applicability, PINNs can be difficult to train, particularly for problems involving strong nonlinearities, multiple scales, or competing physical constraints \cite{krishnapriyan2021characterizing}. Standard formulations may converge slowly or unstably, show strong sensitivity to hyperparameter choices, and produce inaccurate solutions even when the training loss appears well behaved. These difficulties are closely related to the interaction between network representation and the optimization dynamics induced by the composite loss \cite{wang2021understanding, wang2022and}. Poor conditioning of the PINN loss may lead to slow convergence and unstable training \cite{urban2025unveiling, wang2024practical}. These difficulties are typically exacerbated in problems involving steep gradients, multiscale structures, or substantial imbalance among residual terms, thereby motivating targeted modifications to the network architecture or optimization strategy \cite{luo2025physics}.

Recent studies have sought to improve PINN training from both optimization and architectural perspectives. On the optimization side, representative approaches include adaptive loss weighting \cite{wang2021understanding, maddu2022inverse, deguchi2023dynamic, niu2025improved}, gradient-enhanced formulations \cite{yu2020gradient, yu2022gradient}, minimax or dual optimization strategies \cite{liu2021dual}, and methods motivated by neural tangent kernel analysis \cite{wang2022and} or loss-landscape diagnostics \cite{song2024loss}. Architectural developments have followed a parallel direction, including attention-based designs \cite{wang2021understanding}, adaptive activation functions \cite{jagtap2020adaptive}, Fourier feature embeddings \cite{tancik2020fourier}, and Kolmogorov–Arnold networks \cite{liu2024kan}. These modifications can improve the representation of multiscale or highly oscillatory solutions and help alleviate spectral bias. However, architecture and optimization are still commonly designed as separate components. This separation leaves their mutual influence on gradient interactions and training dynamics largely unaccounted for, even though both directly shape the optimization process of PINNs.

The connection between architecture and optimization becomes clearer when PINN training is examined through its composite loss. A standard PINN minimizes a weighted objective composed of several physical constraints, such as PDE residuals and boundary or initial conditions. Each term generates its own gradient signal, and the interaction among these signals directly affects convergence and training stability. Previous studies have shown that gradients associated with different constraints may differ substantially in magnitude or point in conflicting directions, leading to inefficient or unstable optimization \cite{yu2020gradient, liu2021conflict, wang2021understanding}. These gradient characteristics are closely tied to the network architecture, which determines how the physical constraints are represented and how their signals propagate through the model. The optimization strategy, in turn, governs how these competing signals are combined during parameter updates. Gradient conflicts in PINNs therefore cannot be viewed purely as an optimization issue; they also depend on the representation induced by the network architecture. This coupling motivates the joint design of architecture and optimization rather than treating them as separate components.

These observations suggest that improving PINN training requires architecture and optimization to be considered jointly. Architectural choices shape the gradient structure encountered during training, whereas the optimization strategy determines how these gradients are handled during parameter updates. Improving only one side may therefore be insufficient: a more expressive network can still be difficult to optimize, while a sophisticated optimizer cannot overcome limitations in the underlying representation. This provides the motivation for an architecture–optimization co-design framework in which both components are developed in a coordinated manner.

{\color{black}In this work, we develop an \emph{Architecture--Conflict-Resolved PINN} (ACR-PINN) that combines architectural adaptation with gradient-conflict handling. The architectural component, termed \emph{Layer-wise Dynamic Adaptation} (LDA), re-encodes the input coordinates at each hidden layer and applies an input-conditioned, feature-wise gate followed by residual fusion. The resulting LDA-based PINN (LDA-PINN) is used to assess the effect of this architectural modification alone. For optimization, we introduce \emph{Gradient-Conflict-Resolved PINN} (GCR-PINN), in which the PDE residual, initial-condition, and boundary-condition losses are treated as separate tasks. Their gradients are processed using a PINN-specific adaptation of Projecting Conflicting Gradients (PCGrad)~\cite{yu2020gradient}. Thus, GCR-PINN refers to the use of this conflict-resolution mechanism for physical loss terms rather than to a new general-purpose gradient projection rule. ACR-PINN integrates the LDA architecture with GCR-PINN while retaining the original physical loss formulation. Controlled experiments are then used to separate the effects of the two components and examine their interaction across the benchmark problems.}

The main contributions of this work are summarized as follows:
\begin{itemize}
\item {\color{black}
We introduce the LDA architecture for PINNs. At each hidden layer, two independently parameterized encoders construct coordinate features directly from the original input. Input-conditioned, feature-wise gates combine the two branches, and the resulting coordinate feature is incorporated into the multilayer perceptron (MLP) backbone through a scaled residual connection. This design allows the coordinate-to-feature mapping to adapt across network depth.
}
\item {\color{black}
We formulate GCR-PINN for the composite physical loss of PINNs. The PDE, initial-condition, and boundary-condition losses are treated as separate optimization tasks, and negatively aligned task gradients are conditionally projected before aggregation, while nonconflicting gradient pairs remain unchanged.
}
\item {\color{black}
We integrate LDA and GCR-PINN into ACR-PINN, thereby coordinating layer-wise coordinate adaptation with conflict-resolved parameter updates. Parameter-matched ablations and gate interventions indicate that the LDA gains cannot be explained by model size alone and that the predictions depend measurably on the learned coordinate-fusion pathway. Gradient-geometry analyses further show how this representation affects task-gradient compatibility and the correction required by GCR-PINN.
}
\item {\color{black}
We evaluate ACR-PINN across seven PDE benchmarks. Among the four core architecture--optimizer configurations, ACR-PINN attains the lowest observed mean across all reported error metrics. Comparisons with representative PINN methods, together with robustness and computational analyses, further characterize the accuracy gains, their persistence under the examined variations, and the associated computational trade-off.
}
\end{itemize}

{\color{black}Section~\ref{sec:Methodology} presents LDA, GCR-PINN, and their integration into ACR-PINN. Section~\ref{sec:Experiments} reports the evaluation protocol, principal results, mechanism analyses, baseline comparisons, robustness assessments, and computational cost. The appendices consolidate baseline specifications, task-resolved diagnostics, benchmark definitions and implementation details, controlled resource measurements, and supplementary robustness analyses. Section~\ref{sec:Conclusion} summarizes the findings and their scope.}

\section{Methodology}\label{sec:Methodology}

\subsection{Physics-Informed Neural Networks}

PINNs approximate solutions to differential equations using neural networks trained under constraints imposed by the governing equations and prescribed physical conditions~\cite{raissi2019physics}. Unlike purely data-driven models, PINNs do not require solution labels at interior collocation points, since automatic differentiation provides the derivatives needed to evaluate the governing differential operators.

For illustration, consider a scalar time-dependent PDE defined on
$\Omega := \Omega_x \times \Omega_t$:
\begin{equation}
\mathcal{N}[u](x,t) = 0,
\quad
(x,t) \in \Omega,
\end{equation}
subject to the initial condition
\begin{equation}
u(x,t_0) = u_0(x),
\quad
x \in \Omega_x,
\end{equation}
and the Dirichlet boundary condition
\begin{equation}
u(x,t) = u_b(x,t),
\quad
(x,t) \in \partial\Omega_x \times \Omega_t,
\end{equation}
where $\mathcal{N}$ denotes a possibly nonlinear differential operator. Steady-state problems, vector-valued systems, derivative constraints, and periodic boundary conditions can be accommodated by omitting irrelevant terms or introducing the corresponding residuals.

Throughout this work, $u$ denotes the analytical or numerical reference solution, whereas $u_\theta$ denotes the neural approximation parameterized by $\theta$. The prescribed initial and boundary data are denoted by $u_0$ and $u_b$, respectively. Subscripts $f$, $i$, and $b$ refer to the interior collocation, initial, and boundary point sets. For example, $u_{\theta,f}$ denotes the values of $u_\theta$ evaluated at the interior collocation points. The network maps the input coordinates to the physical field of interest, while automatic differentiation provides the derivatives required to construct the governing residuals.

\begin{figure}[H]
\centering
\includegraphics[width=0.85\textwidth]{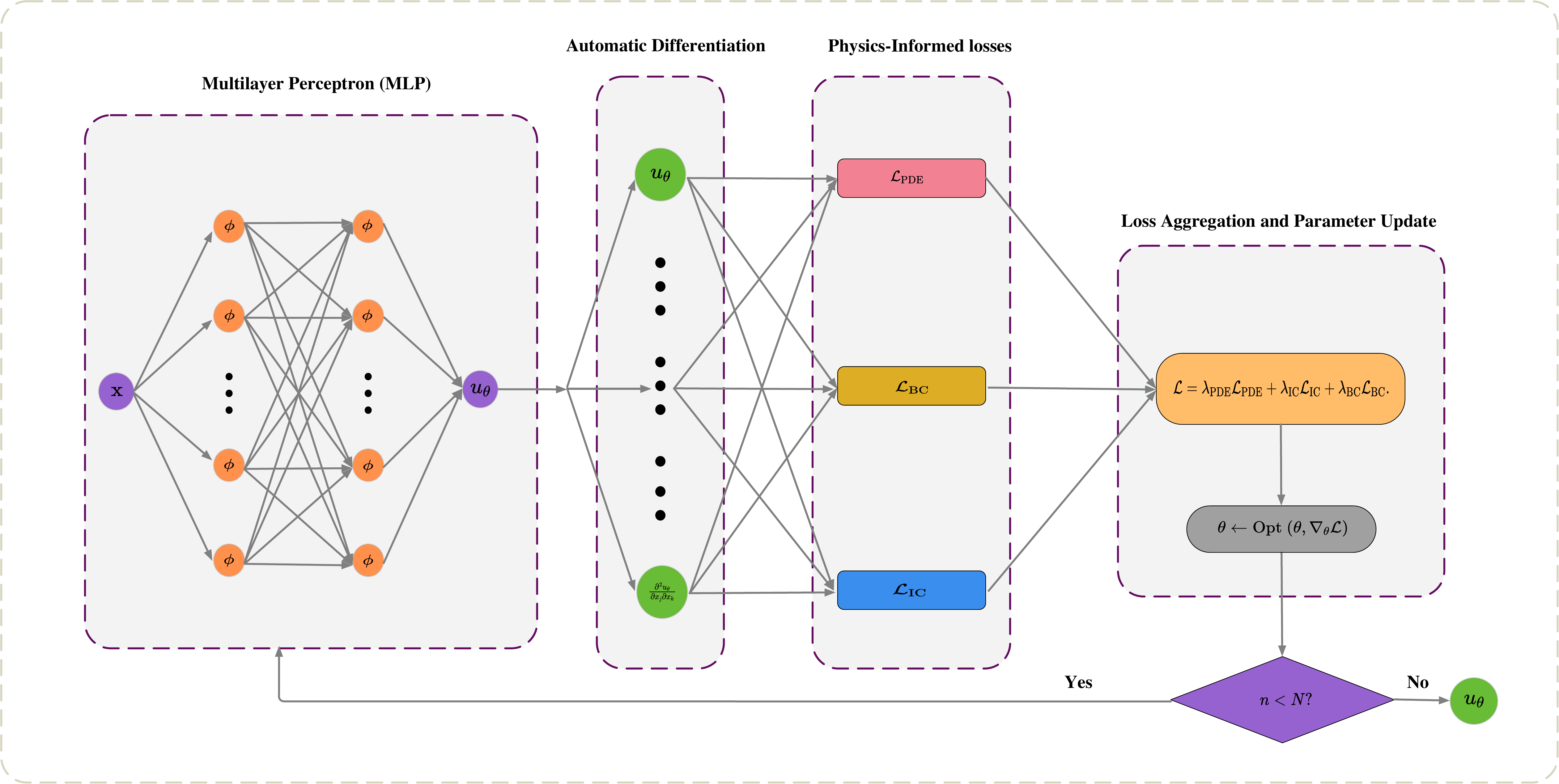}
\caption{Standard PINN framework. The neural approximation $u_\theta$ and the derivatives obtained by automatic differentiation are used to construct the physical losses. The PDE, initial-condition, and boundary-condition terms are then combined through a weighted sum to update the network parameters.}
\label{fig:pinn_framework}
\end{figure}

As illustrated in Fig.~\ref{fig:pinn_framework}, the standard PINN baseline (Std-PINN) is trained by minimizing a weighted combination of the PDE residual, initial-condition (IC), and boundary-condition (BC) losses:
\begin{equation}
\mathcal{L}(\theta)
=
\lambda_{\mathrm{PDE}}\mathcal{L}_{\mathrm{PDE}}
+
\lambda_{\mathrm{IC}}\mathcal{L}_{\mathrm{IC}}
+
\lambda_{\mathrm{BC}}\mathcal{L}_{\mathrm{BC}},
\label{eq:total_loss}
\end{equation}
where $\lambda_{\mathrm{PDE}}$, $\lambda_{\mathrm{IC}}$, and $\lambda_{\mathrm{BC}}$ control the relative contributions of the three terms.

For sampled interior collocation, initial, and boundary points, the corresponding losses are defined as
\begin{align}
\mathcal{L}_{\mathrm{PDE}}
&=
\frac{1}{N_f}
\sum_{i=1}^{N_f}
\left\|
\mathcal{N}\!\left[u_\theta(x_f^{(i)},t_f^{(i)})\right]
\right\|_2^2, \\
\mathcal{L}_{\mathrm{IC}}
&=
\frac{1}{N_i}
\sum_{i=1}^{N_i}
\left\|
u_\theta(x_i^{(i)},t_0)-u_0(x_i^{(i)})
\right\|_2^2, \\
\mathcal{L}_{\mathrm{BC}}
&=
\frac{1}{N_b}
\sum_{i=1}^{N_b}
\left\|
u_\theta(x_b^{(i)},t_b^{(i)})
-
u_b(x_b^{(i)},t_b^{(i)})
\right\|_2^2.
\end{align}
Terms that are not relevant to a particular problem are omitted, while additional physical constraints can be incorporated through analogous residual terms.

Minimizing Eq.~\eqref{eq:total_loss} enforces the governing equation and prescribed conditions through soft penalties at the sampled points. For multiscale or highly oscillatory problems, however, the resulting representation can be difficult to optimize, while gradients associated with different physical constraints may differ substantially in magnitude or direction~\cite{krishnapriyan2021characterizing,wang2021understanding}. The following subsections address these two aspects through layer-wise coordinate adaptation and gradient conflict resolution.

\subsection{\textcolor{black}{Layer-wise Dynamic Adaptation via Feature-wise Gating}}

The LDA architecture extends a standard MLP by introducing two independently parameterized coordinate encoders at each hidden layer together with an input-conditioned, feature-wise gating mechanism. The gate adaptively combines the two encoded branches for each feature channel, making the coordinate representation dependent on both the layer and the input. Although LDA is inspired by the adaptive weighting principle underlying attention mechanisms~\cite{vaswani2017attention}, it does not adopt the Transformer-style self-attention formulation. Instead, it performs input-conditioned, feature-wise gating between two coordinate-encoding branches.

\subsubsection{\textcolor{black}{Backbone and Layer-wise Coordinate Features}}

Let $\mathbf{x}\in\mathbb{R}^{d}$ denote the complete coordinate vector; for a time-dependent problem, $\mathbf{x}$ includes both spatial and temporal coordinates. A standard MLP with $L$ hidden layers computes
\begin{equation}
a_\ell(\mathbf{x})
=
\phi\!\left(W_\ell a_{\ell-1}(\mathbf{x}) + b_\ell\right),
\quad
a_0(\mathbf{x})=\mathbf{x},
\quad
\ell=1,\ldots,L,
\end{equation}
where $a_\ell\in\mathbb{R}^{h_\ell}$, $W_\ell$ and $b_\ell$ are trainable parameters, and $\phi$ denotes the hidden-layer activation function.

In a standard MLP, deeper representations are constructed recursively from the preceding hidden features, and the original coordinates are directly accessed only at the input layer. LDA extends this structure by introducing a separate coordinate pathway at each hidden layer, so that layer-specific coordinate features can be constructed directly from the original input at different network depths.

\subsubsection{Two-Branch Coordinate Re-encoding}

At each hidden layer $\ell$, two layer-specific encoders map the original coordinate vector $\mathbf{x}$ to distinct coordinate representations,
$e_\ell^{(1)}(\mathbf{x})$ and $e_\ell^{(2)}(\mathbf{x})$.
Unlike a shared input encoding reused throughout the network, these representations are constructed independently at each depth.

Specifically, the two coordinate encodings at layer $\ell$ are defined as
\begin{subequations}
\begin{align}
e_\ell^{(1)}(\mathbf{x})
&=
\phi\!\left(E_\ell^{(1)}\mathbf{x}+c_\ell^{(1)}\right),\\
e_\ell^{(2)}(\mathbf{x})
&=
\phi\!\left(E_\ell^{(2)}\mathbf{x}+c_\ell^{(2)}\right),
\end{align}
\end{subequations}
where
$E_\ell^{(1)},E_\ell^{(2)}\in\mathbb{R}^{h_\ell\times d}$
and
$c_\ell^{(1)},c_\ell^{(2)}\in\mathbb{R}^{h_\ell}$
are independently learned parameters. Both branches therefore produce
$h_\ell$-dimensional features, matching the width of the corresponding hidden layer.

This construction is related to the modified input-encoding architecture of Wang et al.~\cite{wang2021understanding}, in which two transformed input representations are reused across network depth. LDA instead assigns independent coordinate encoders to each hidden layer, allowing the encoded coordinate features to vary with depth. No predefined role is imposed on either branch; their relative contributions are determined by the feature-gating mechanism introduced in the next subsection.

\subsubsection{\textcolor{black}{Layer-wise Dynamic Feature-Gating Mechanism}}

At each hidden layer, the gating module assigns input-dependent weights to the two coordinate-encoding branches for each feature channel. Let
\begin{equation}
a_\ell^{\mathrm{MLP}}(\mathbf{x})
=
\phi\!\left(W_\ell a_{\ell-1}(\mathbf{x}) + b_\ell\right)
\end{equation}
denote the intermediate backbone representation at layer $\ell$. This representation is concatenated with the two coordinate features:
\begin{equation}
z_\ell(\mathbf{x})
=
\left[
a_\ell^{\mathrm{MLP}}(\mathbf{x});
e_\ell^{(1)}(\mathbf{x});
e_\ell^{(2)}(\mathbf{x})
\right]
\in\mathbb{R}^{3h_\ell}.
\end{equation}

A layer-specific gating network with one hidden layer of width $h_\ell$ maps
$z_\ell(\mathbf{x})$ to $2h_\ell$ logits:
\begin{equation}
g_\ell(z)
=
W_{\ell,2}^{g}
\phi\!\left(W_{\ell,1}^{g}z+b_{\ell,1}^{g}\right)
+b_{\ell,2}^{g}
\in\mathbb{R}^{2h_\ell}.
\end{equation}
These logits are reshaped as
\begin{equation}
G_\ell(\mathbf{x})
=
\operatorname{reshape}_{2\times h_\ell}
\!\left(g_\ell(z_\ell(\mathbf{x}))\right)
\in\mathbb{R}^{2\times h_\ell},
\end{equation}
where the first dimension corresponds to the two encoding branches and the second to the $h_\ell$ feature channels.

For each channel $k$, a softmax over the two branches gives
\begin{equation}
\alpha_{\ell,k}^{(i)}(\mathbf{x})
=
\frac{
\exp\!\left(G_{\ell,i,k}(\mathbf{x})\right)
}{
\sum_{j=1}^{2}
\exp\!\left(G_{\ell,j,k}(\mathbf{x})\right)
},
\quad
i=1,2,
\qquad
k=1,\dots,h_\ell.
\end{equation}
Hence,
$\alpha_{\ell,k}^{(1)}(\mathbf{x})+\alpha_{\ell,k}^{(2)}(\mathbf{x})=1$
for every feature channel. Because the gate is conditioned jointly on the current backbone representation and the two coordinate features, the relative contribution of each branch can vary across inputs, layers, and feature channels.

\subsubsection{Feature-wise Gated Fusion with Residual Injection}

The two coordinate representations are fused using the learned feature-wise gating weights:
\begin{equation}
m_{\ell,k}(\mathbf{x})
=
\alpha_{\ell,k}^{(1)}(\mathbf{x})e_{\ell,k}^{(1)}(\mathbf{x})
+
\alpha_{\ell,k}^{(2)}(\mathbf{x})e_{\ell,k}^{(2)}(\mathbf{x}),
\quad
k=1,\dots,h_\ell,
\end{equation}
where
$m_\ell(\mathbf{x})\in\mathbb{R}^{h_\ell}$
denotes the fused coordinate modulation.

This modulation is incorporated into the backbone through a scaled residual connection:
\begin{equation}
a_\ell(\mathbf{x})
=
a_\ell^{\mathrm{MLP}}(\mathbf{x})
+
\rho_\ell m_\ell(\mathbf{x}),
\end{equation}
where $\rho_\ell$ controls the strength of the coordinate contribution at layer $\ell$.

After the final hidden layer, the neural approximation is obtained through a linear output map:
\begin{equation}
u_\theta(\mathbf{x})
=
W_{\mathrm{out}}a_L(\mathbf{x})
+
b_{\mathrm{out}}.
\end{equation}

\begin{figure}[H]
\centering
\includegraphics[width=0.85\textwidth]{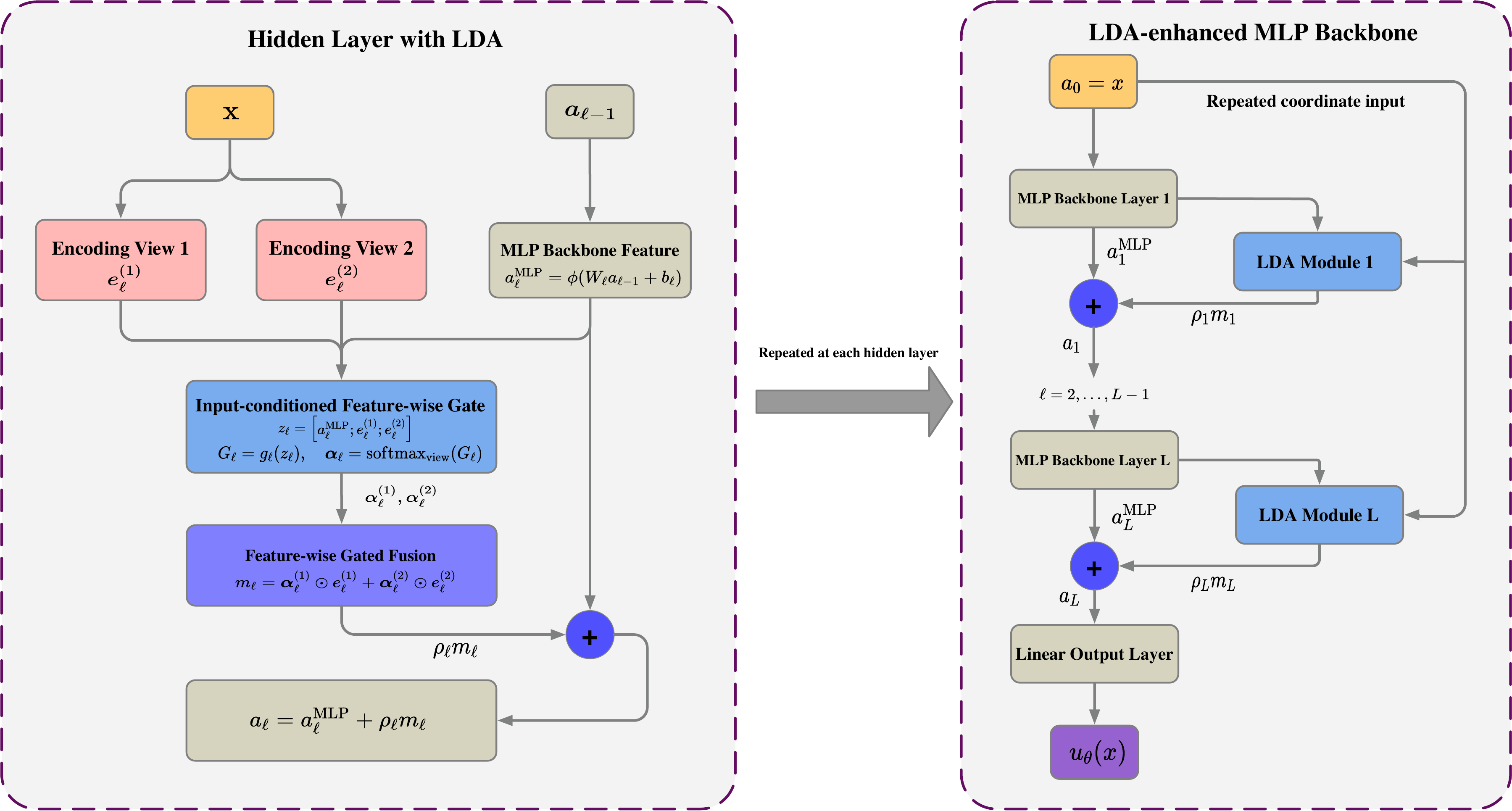}
\caption{\textcolor{black}{
Layer-wise dynamic adaptation through feature-wise gated fusion. At hidden layer $\ell$, input-conditioned weights combine two coordinate representations into $m_\ell(\mathbf{x})$, and the scaled term $\rho_\ell m_\ell(\mathbf{x})$ is added to the backbone feature $a_\ell^{\mathrm{MLP}}(\mathbf{x})$. The same residual adaptation is repeated across the hidden layers.
}}
\label{fig:LDAframework}
\end{figure}

Figure~\ref{fig:LDAframework} summarizes the resulting architecture. The left panel illustrates the gated fusion at a single hidden layer, while the right panel shows how the same adaptation mechanism is repeated throughout the MLP backbone.

For the LDA configuration used in the main experiments and controlled comparisons, $\rho_\ell$ is trainable and initialized to $0.1$. The final affine layer of $g_\ell$ is initialized to zero, yielding
$\alpha_{\ell,k}^{(1)}(\mathbf{x})
=
\alpha_{\ell,k}^{(2)}(\mathbf{x})
=
0.5$
at the start of training. The backbone, coordinate encoders, gating networks, and residual scales are optimized jointly.

\begin{figure}[H]
\centering
\includegraphics[width=\textwidth]{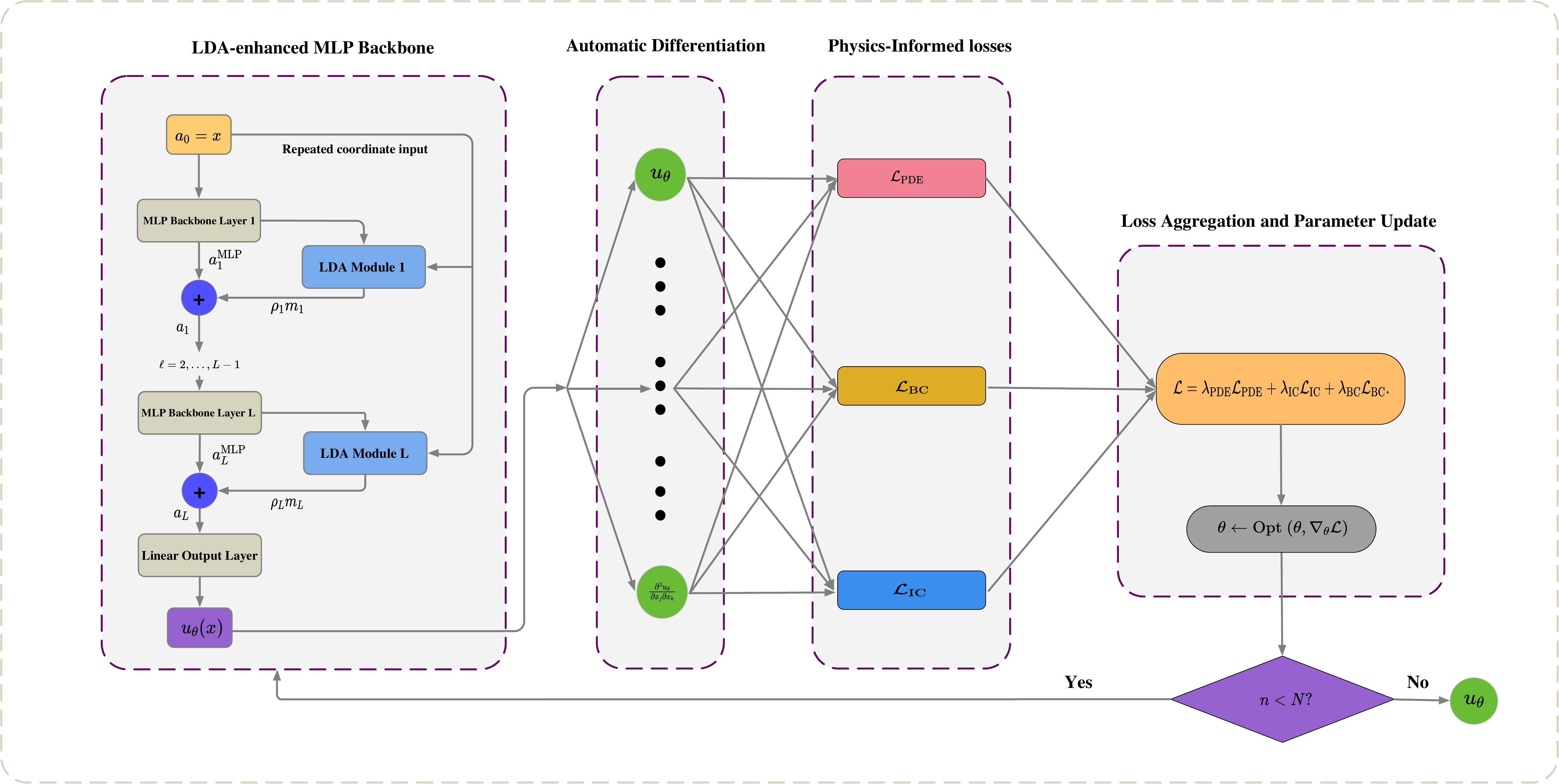}
\caption{\textcolor{black}{
Training framework of LDA-PINN. The LDA backbone maps the input coordinates to $u_\theta(\mathbf{x})$, while automatic differentiation provides the derivatives required to evaluate the PDE residual. The PDE, initial-condition, and boundary-condition losses are then combined through the standard weighted-sum formulation.
}}
\label{fig:LDA_PINN_framework}
\end{figure}

Embedding the LDA backbone into the physics-informed formulation yields LDA-PINN, as shown in Fig.~\ref{fig:LDA_PINN_framework}. Relative to Std-PINN, LDA-PINN changes only the network parameterization while retaining the same physical loss terms and standard weighted-sum optimization.

\FloatBarrier
\subsection{\textcolor{black}{Gradient-Conflict-Resolved PINN}}

{\color{black}The network architecture determines how the physical solution is represented, whereas the optimization rule determines how the individual physical constraints update the shared parameters. Because these constraints can generate competing gradient directions, we complement the LDA backbone with a conflict-resolved update for the component losses.}

\subsubsection{\textcolor{black}{Gradient Conflicts in Multi-Task PINN Training}}

PINN training can be viewed as a multi-task optimization problem because several physical constraints are enforced through the same network parameters. A standard formulation typically includes the PDE residual, IC, and BC losses. Each component defines a separate objective, but all are optimized with respect to the shared parameter vector $\theta$.

The gradient associated with task $i$ is
\begin{equation}
\mathbf{g}_i
=
\nabla_{\theta}\mathcal{L}_i(\theta),
\end{equation}
where $\mathcal{L}_i$ denotes the corresponding loss component. If the task gradients are aligned, their joint update can reduce several objectives simultaneously. In PINNs, however, the physical constraints generally encode different aspects of the governing problem and are evaluated on different point sets. Their gradients may therefore differ in both magnitude and direction during training.

A gradient conflict between tasks $i$ and $j$ is identified when
\begin{equation}
\mathbf{g}_i^{\top}\mathbf{g}_j < 0,
\label{eq:gradient_conflict}
\end{equation}
which means that the two gradients form an obtuse angle. In this case, a descent direction favored by one task can oppose that favored by the other, so direct summation may introduce destructive interference. Figure~\ref{fig:gradient_conflict_geometry} illustrates the geometric difference between cooperative and conflicting gradients.

\begin{figure}[H]
\centering
\begin{tikzpicture}[
>=Stealth,
scale=0.95,
gi/.style={->, very thick, red!80!black},
gj/.style={->, very thick, teal!70!black},
ang/.style={dashed, thick}
]
\def\L{2.8}
\def\R{0.75}
\def\theta{40}

\begin{scope}[xshift=-3.6cm]
\draw[gi] (0,0) -- (\L,0) node[below right] {$\mathbf{g}_i$};
\draw[gj] (0,0) -- ({\L*cos(\theta)}, {\L*sin(\theta)})
node[above] {$\mathbf{g}_j$};
\draw[ang] (\R,0) arc[start angle=0, end angle=\theta, radius=\R];
\node at (\L/2,-0.9) {\small (a) $\cos\phi>0$};
\end{scope}

\begin{scope}[xshift=3.6cm]
\draw[gi] (0,0) -- (\L,0) node[below right] {$\mathbf{g}_i$};
\draw[gj] (0,0) -- ({-\L*cos(\theta)}, {\L*sin(\theta)})
node[above left] {$\mathbf{g}_j$};
\draw[ang] (\R,0) arc[start angle=0, end angle=180-\theta, radius=\R];
\node at (\L/2,-0.9) {\small (b) $\cos\phi<0$};
\end{scope}
\end{tikzpicture}
\caption{Pairwise task-gradient geometry in PINN optimization. A positive cosine indicates locally cooperative descent directions, whereas a negative cosine identifies a directional conflict.}
\label{fig:gradient_conflict_geometry}
\end{figure}

Such conflicts can arise repeatedly during PINN training, particularly when the physical losses evolve at different rates or the target solution contains multiscale or high-frequency features. Under these conditions, simple scalar aggregation may lead to slow convergence, unstable updates, or uneven satisfaction of the physical constraints. This motivates an optimization rule that acts directly on the task gradients when directional conflicts occur.

\subsubsection{\textcolor{black}{Conditional Pairwise Gradient Projection}}

Adaptive loss reweighting is commonly used in PINNs to reduce imbalance among physical constraints. Such methods can adjust differences in gradient magnitude by rescaling the corresponding loss terms, but they do not directly resolve directional conflicts between task gradients. When two gradients point toward opposing descent directions, scalar reweighting alone cannot remove the conflicting component. This motivates an update rule that operates directly in gradient space.

{\color{black}We apply the PCGrad algorithm of Yu et al.~\cite{yu2020gradient} to PINN optimization by treating the PDE residual, initial-condition, and boundary-condition losses as separate tasks. We refer to this PINN-specific application as GCR-PINN. Conditional pairwise projection is performed before the adjusted task gradients are aggregated to update the shared parameters.}

For two loss components $\mathcal{L}_i$ and $\mathcal{L}_j$, define
\begin{equation}
\mathbf{g}_i=\nabla_{\theta}\mathcal{L}_i(\theta),
\qquad
\mathbf{g}_j=\nabla_{\theta}\mathcal{L}_j(\theta)
\end{equation}
denote the gradients of two PINN loss components. A conflict is detected when
\begin{equation}
\mathbf{g}_i^{\top}\mathbf{g}_j < 0.
\end{equation}
In this case, $\mathbf{g}_i$ is projected onto the normal plane of $\mathbf{g}_j$:
\begin{equation}
\mathbf{g}_i
\leftarrow
\mathbf{g}_i
-
\frac{\mathbf{g}_i^{\top}\mathbf{g}_j}
{\|\mathbf{g}_j\|_2^2}
\mathbf{g}_j.
\end{equation}
This update removes the component of $\mathbf{g}_i$ that conflicts with $\mathbf{g}_j$ while preserving the remaining descent direction. Figure~\ref{fig:gcr_two_projections} illustrates the corresponding geometry.

\begin{figure}[htbp]
\centering
\begin{tikzpicture}[
>=Stealth,
scale=0.95,
gi/.style={->, very thick, red!80!black},
gj/.style={->, very thick, teal!70!black},
gproj/.style={->, very thick, dashed},
arcproj/.style={dashed, thick}
]
\def\L{2.8}
\def\R{0.85}
\def\theta{40}

\pgfmathsetmacro{\angGj}{180-\theta}
\pgfmathsetmacro{\angGjProj}{90}
\pgfmathsetmacro{\angGi}{0}
\pgfmathsetmacro{\angGiProj}{\theta}

\begin{scope}[xshift=-4.0cm]
\draw[gi] (0,0) -- (\L,0) node[below right] {$\mathbf{g}_i$};
\draw[gj] (0,0) -- ({\L*cos(\angGj)},{\L*sin(\angGj)})
node[above left] {$\mathbf{g}_j$};

\draw[gproj, teal!70!black]
(0,0) -- (0,{\L*sin(\angGj)})
node[right] {$\mathbf{g}_j^{\perp i}$};

\draw[arcproj]
({\R*cos(\angGj)},{\R*sin(\angGj)})
arc[start angle=\angGj, end angle=\angGjProj, radius=\R];

\node at (\L/2,-1.0)
{\small \textbf{(a)} Project $\mathbf{g}_j$ onto $\mathbf{n}_i$};
\end{scope}

\begin{scope}[xshift=4.0cm]
\draw[gi] (0,0) -- (\L,0) node[below right] {$\mathbf{g}_i$};
\draw[gj] (0,0) -- ({\L*cos(\angGj)},{\L*sin(\angGj)})
node[above left] {$\mathbf{g}_j$};

\draw[gproj, red!80!black]
(0,0) --
({\L*cos(\theta)*cos(\theta)},
 {\L*cos(\theta)*sin(\theta)})
node[above right] {$\mathbf{g}_i^{\perp j}$};

\draw[arcproj]
({\R*cos(\angGi)},{\R*sin(\angGi)})
arc[start angle=\angGi, end angle=\angGiProj, radius=\R];

\node at (\L/2,-1.0)
{\small \textbf{(b)} Project $\mathbf{g}_i$ onto $\mathbf{n}_j$};
\end{scope}
\end{tikzpicture}
\caption{Conditional pairwise projection in GCR-PINN. For a conflicting gradient pair, the projection removes the component of one task gradient that opposes the other.}
\label{fig:gcr_two_projections}
\end{figure}

For multiple tasks, the pairwise projections are applied in randomized order, and the adjusted gradients are then aggregated to form the final update. If
$\mathbf{g}_i^{\top}\mathbf{g}_j \ge 0$,
the corresponding pair is left unchanged. GCR-PINN therefore modifies only conflicting directions while preserving cooperative gradient interactions.

\begin{algorithm}[htbp]
\caption{\textcolor{black}{Conditional pairwise gradient update in GCR-PINN}}
\label{alg:gcr_pinn_update}
\begin{algorithmic}[1]
\Require
Task losses $\{\mathcal{L}_k(\theta)\}_{k=1}^{K}$, shared parameters $\theta$
\Ensure
Conflict-resolved gradient $\mathbf{g}^{\mathrm{CR}}$

\State Compute task gradients
$\mathbf{g}_k \gets \nabla_{\theta}\mathcal{L}_k(\theta)$,
$k=1,\dots,K$

\State Initialize
$\tilde{\mathbf{g}}_k \gets \mathbf{g}_k$

\For{$i=1$ to $K$}
    \State Draw a random permutation $\pi$ of $\{1,\dots,K\}\setminus\{i\}$
    \For{each $j\in\pi$}
        \If{$\tilde{\mathbf{g}}_i^{\top}\mathbf{g}_j<0$}
            \State
            $\tilde{\mathbf{g}}_i \gets
            \tilde{\mathbf{g}}_i -
            \frac{\tilde{\mathbf{g}}_i^{\top}\mathbf{g}_j}
            {\|\mathbf{g}_j\|_2^2}\mathbf{g}_j$
        \EndIf
    \EndFor
\EndFor

\State Aggregate
$\mathbf{g}^{\mathrm{CR}}
\gets
\sum_{k=1}^{K}\tilde{\mathbf{g}}_k$

\State \Return $\mathbf{g}^{\mathrm{CR}}$
\end{algorithmic}
\end{algorithm}

{\color{black}Algorithm~\ref{alg:gcr_pinn_update} summarizes the GCR-PINN update. The procedure acts only on gradient pairs with negative inner products, while cooperative pairs are retained. This directional correction is complementary to scalar loss weighting, since the two mechanisms act on different properties of the task gradients.}

\subsubsection{\textcolor{black}{Integration into PINN Training}}

GCR-PINN modifies the gradient aggregation stage of standard PINN training while leaving the network architecture and physical loss formulation unchanged. The PDE residual, initial-condition, and boundary-condition losses are treated as separate optimization tasks, and their gradients are computed individually with respect to the shared parameters. Pairwise conflicts are then resolved before the adjusted gradients are aggregated to form the parameter update.

\begin{figure}[H]
\centering
\includegraphics[width=\textwidth]{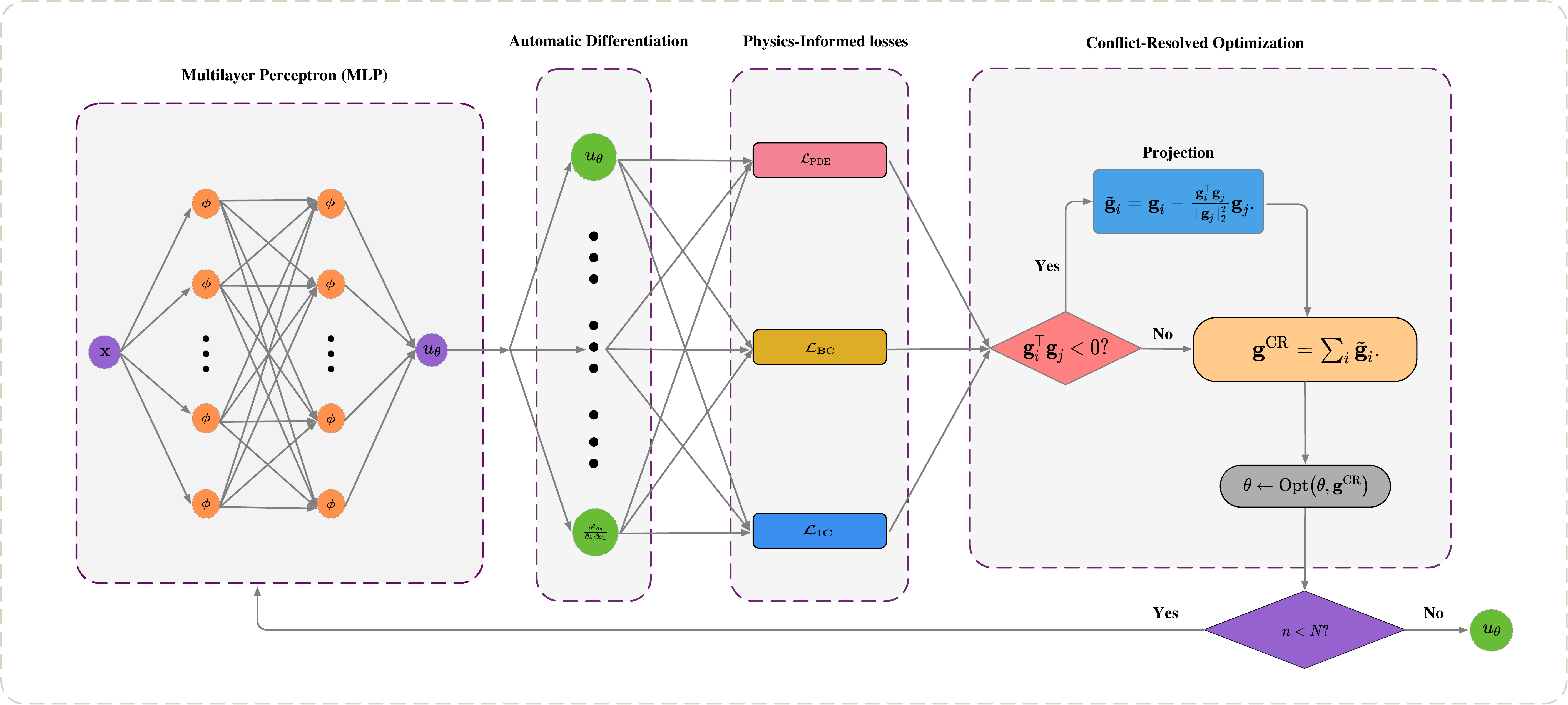}
\caption{\textcolor{black}{
Training framework of GCR-PINN. The PDE residual, initial-condition, and boundary-condition losses are treated as separate tasks and generate their corresponding gradients. Conflicting gradient pairs satisfying $\mathbf{g}_i^\top\mathbf{g}_j<0$ undergo conditional projection, whereas nonconflicting pairs are left unchanged. The adjusted gradients are aggregated into $\mathbf{g}^{\mathrm{CR}}$ and passed to the optimizer to update the shared parameters.
}}
\label{fig:GCR_PINN_framework}
\end{figure}

As illustrated in Fig.~\ref{fig:GCR_PINN_framework}, the main difference from Std-PINN lies in the gradient-processing step between loss evaluation and parameter update. Gradient pairs with negative inner products undergo the conditional projection described above, whereas nonconflicting pairs remain unchanged. The resulting task gradients are aggregated into the conflict-resolved direction $\mathbf{g}^{\mathrm{CR}}$, which is passed to the underlying optimizer to update the shared parameters.

{\color{black}
GCR-PINN thus provides an optimization-side modification that is independent of the network architecture. This separation allows its effect to be evaluated alone and, subsequently, in combination with the LDA architecture.
}

\subsection{\textcolor{black}{Architecture--Optimization Co-Design}}

The LDA architecture and GCR-PINN operate at different stages of PINN training. LDA provides layer-wise coordinate adaptation by modifying the coordinate-to-feature representation used to evaluate the physical losses, whereas GCR-PINN performs gradient conflict resolution before the resulting task gradients are aggregated to update the shared parameters. Because task-gradient geometry depends on the underlying network representation, the two components can be integrated within a common training framework.

{\color{black}The resulting ACR-PINN method uses the LDA backbone to construct the neural approximation and applies GCR-PINN to the gradients associated with the PDE residual, initial-condition, and boundary-condition losses, while leaving the physical loss formulation unchanged.}

\begin{figure}[H]
\centering
\includegraphics[width=\textwidth]{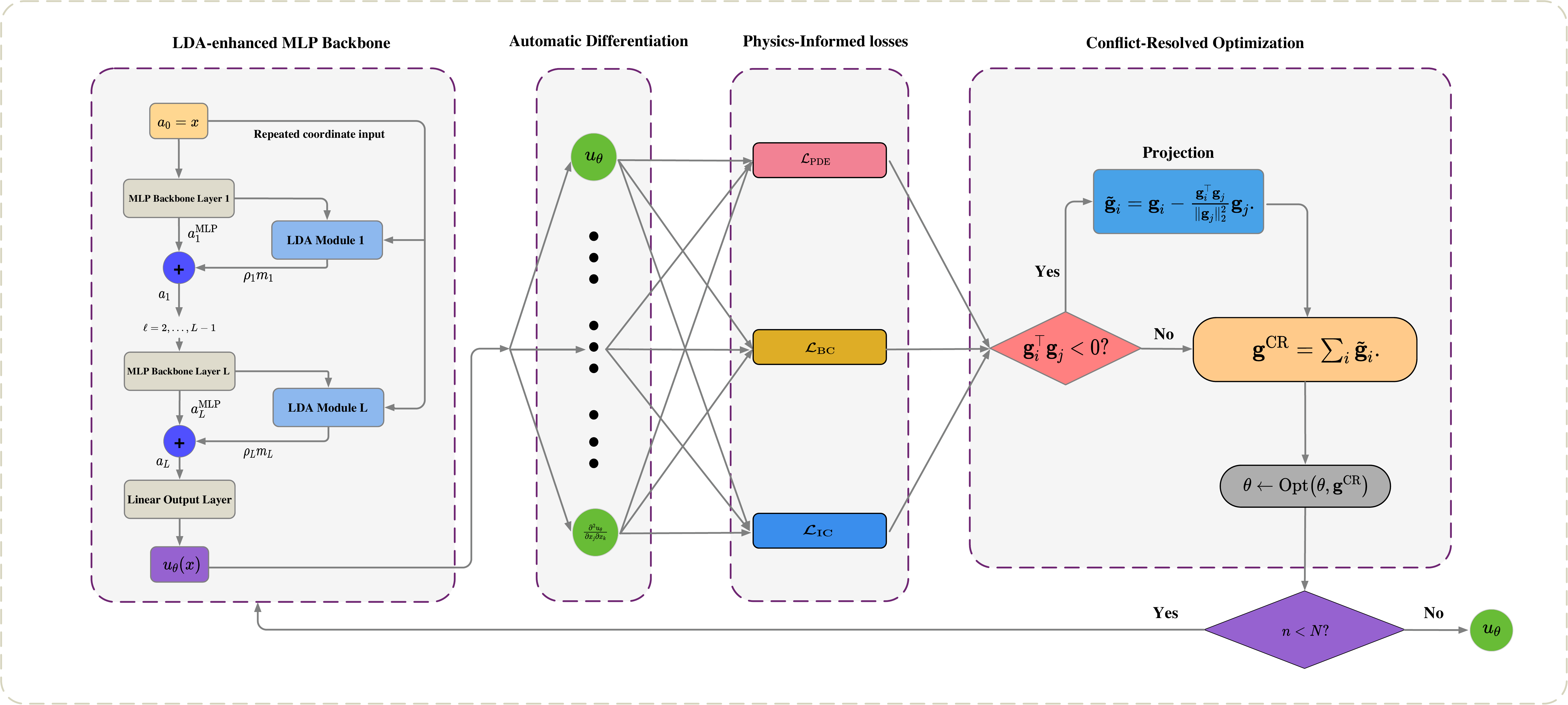}
\caption{\textcolor{black}{
Architecture--optimization framework of ACR-PINN. The LDA backbone maps the input coordinates to the neural representation, from which the physical losses generate task-specific gradients
$\mathbf{g}_i=\nabla_{\theta}\mathcal{L}_i$.
GCR-PINN conditionally projects negatively aligned gradient pairs and aggregates the adjusted gradients as $\mathbf{g}^{\mathrm{CR}}$ to update the shared LDA parameters.
}}
\label{fig:ACR_PINN_framework}
\end{figure}

As shown in Fig.~\ref{fig:ACR_PINN_framework}, the physical losses evaluated from the LDA representation generate task-specific gradients
\begin{equation}
\mathbf{g}_i=\nabla_{\theta}\mathcal{L}_i,
\end{equation}
which are processed by GCR-PINN before aggregation. Gradient pairs with negative inner products undergo the conditional projection described above, and the adjusted gradients are then combined to update the shared LDA parameters.

\FloatBarrier
The ACR-PINN training procedure is summarized in Algorithm~\ref{alg:acr_pinn}. Algorithm~\ref{alg:acr_pinn} combines the LDA forward model with the GCR-PINN gradient update in a single training loop. The physical losses are evaluated separately, their gradients are conflict-resolved before aggregation, and the resulting direction is passed to the underlying optimizer to update the shared parameters.

\begin{algorithm}[H]
\caption{Training procedure for ACR-PINN}
\label{alg:acr_pinn}
\begin{algorithmic}[1]
\Require
Training sets $\mathcal{D}_f,\mathcal{D}_{ic},\mathcal{D}_{bc}$;
LDA-network parameters $\theta$;
optimizer $\mathrm{Opt}(\cdot)$ with step size $\eta$;
number of optimization iterations $N$
\Ensure
Trained parameters $\theta$

\For{$n=1$ to $N$}
    \State Sample interior collocation points $(x_f,t_f)\sim\mathcal{D}_f$
    \State Sample initial points $(x_i,t_0)\sim\mathcal{D}_{ic}$ and boundary points $(x_b,t_b)\sim\mathcal{D}_{bc}$

    \State Evaluate the LDA network:
    \State \hspace{1em}
    $u_{\theta,f}\gets u_\theta(x_f,t_f)$,
    $u_{\theta,i}\gets u_\theta(x_i,t_0)$,
    $u_{\theta,b}\gets u_\theta(x_b,t_b)$

    \State Compute the physical losses:
    \State \hspace{1em}
    $\mathcal{L}_{\mathrm{PDE}}\gets\frac{1}{N_f}\sum\left\|\mathcal{N}[u_{\theta,f}]\right\|_2^2$
    \State \hspace{1em}
    $\mathcal{L}_{\mathrm{IC}}\gets\frac{1}{N_i}\sum\left\|u_{\theta,i}-u_0\right\|_2^2$
    \State \hspace{1em}
    $\mathcal{L}_{\mathrm{BC}}\gets\frac{1}{N_b}\sum\left\|u_{\theta,b}-u_b\right\|_2^2$

    \State Compute task gradients:
    \State \hspace{1em}
    $\mathbf{g}_1 \gets \nabla_\theta \mathcal{L}_{\mathrm{PDE}}$,
    $\mathbf{g}_2 \gets \nabla_\theta \mathcal{L}_{\mathrm{IC}}$,
    $\mathbf{g}_3 \gets \nabla_\theta \mathcal{L}_{\mathrm{BC}}$
    \State Initialize $\tilde{\mathbf{g}}_k\gets\mathbf{g}_k$, $k\in\{1,2,3\}$

    \For{$i \in \{1,2,3\}$}
        \State Draw a random order $\pi$ of $\{1,2,3\}\setminus\{i\}$
        \For{each $j\in\pi$}
            \If{$\tilde{\mathbf{g}}_i^\top\mathbf{g}_j<0$}
                \State
                $\tilde{\mathbf{g}}_i\gets\tilde{\mathbf{g}}_i-
                \frac{\tilde{\mathbf{g}}_i^\top\mathbf{g}_j}{\|\mathbf{g}_j\|_2^2}\mathbf{g}_j$
            \EndIf
        \EndFor
    \EndFor

    \State Aggregate the adjusted gradients:
    \State \hspace{1em}
    $\mathbf{g}^{\mathrm{CR}}\gets\tilde{\mathbf{g}}_1+\tilde{\mathbf{g}}_2+\tilde{\mathbf{g}}_3$

    \State Update the shared parameters:
    \State \hspace{1em}
    $\theta\gets\mathrm{Opt}(\theta;\mathbf{g}^{\mathrm{CR}},\eta)$
\EndFor
\end{algorithmic}
\end{algorithm}

\clearpage
\section{Numerical Experiments}\label{sec:Experiments}

\subsection{\textcolor{black}{Evaluation Framework}}
\label{sec:experiment_protocol}
{\color{black}
The numerical study examines three questions: whether the proposed formulation improves predictive accuracy across distinct classes of PDEs, how the architectural and optimization components contribute to that behavior, and whether the resulting method remains competitive and robust under controlled comparisons. Seven benchmark cases are considered, spanning transient and stationary equations, nonlinear and oscillatory solutions, fluid flow, and the five-dimensional (5D) Poisson problem. Their governing equations, domains, and physical constraints are specified in~\ref{app:benchmark_definitions}.

To distinguish the effects of representation and gradient treatment, the core comparison follows a two-by-two design: the MLP and LDA backbones are each trained using direct gradient summation and GCR-PINN. This yields Std-PINN, LDA-PINN, GCR-PINN, and ACR-PINN. Within each benchmark, the four conditions share the sampling strategy, optimization schedule, training budget, and evaluation set. Each condition is evaluated in five paired runs, and prediction errors are computed only after the prescribed training budget on points excluded from both training and model selection. Full sampling, network, and evaluation specifications are provided in~\ref{app:network_evaluation}.
}

For a common evaluation set $\{\mathbf{x}_j\}_{j=1}^{N}$, predictive accuracy is measured by the relative $L_2$ error, relative $L_\infty$ error, and mean-squared error (MSE):
\begin{subequations}
\begin{gather}
E_2
=
\frac{\left(\sum_{j=1}^{N}|u_{\theta,j}-u_j|^2\right)^{1/2}}
{\left(\sum_{j=1}^{N}|u_j|^2\right)^{1/2}},\\
E_\infty
=
\frac{\max_j |u_{\theta,j}-u_j|}{\max_j |u_j|},\\
\mathrm{MSE}
=
\frac{1}{N}\sum_{j=1}^{N}|u_{\theta,j}-u_j|^2.
\end{gather}
\end{subequations}

{\color{black}
Here, $E_2$ measures the global relative error, $E_\infty$ captures the largest normalized pointwise deviation, and MSE retains the absolute error scale. The metrics are applied to the scalar solution for Burgers, Helmholtz, Klein--Gordon, and Poisson, to the velocity magnitude for the lid-driven cavity, and to the complex-valued solution for nonlinear Schr\"odinger. Problem-specific evaluation conventions are detailed in~\ref{app:benchmark_definitions}.
}

\subsection{\textcolor{black}{Overall Performance}}
\label{sec:core_seven}
\subsubsection{\textcolor{black}{Quantitative Comparison across Seven Benchmark Cases}}
{\color{black}
We first examine whether the gains persist across the benchmark suite and how the architectural and optimization components contribute across different problems. Relative $L_2$ error characterizes domain-wide solution accuracy, relative $L_\infty$ error captures the largest local discrepancy, and MSE provides a scale-sensitive measure of the squared approximation error.

{\small\setlength{\tabcolsep}{3pt}
\begin{table}[htbp]
\centering\color{black}\captionsetup{font={color=black}}
\caption{Relative $L_2$ errors for the seven benchmark cases, reported as mean $\pm$ sample standard deviation over five paired runs. The method with the lowest mean error in each row is highlighted in bold.}
\label{tab:s17_core_dynamic}
\resizebox{\textwidth}{!}{\begin{tabular}{lcccc}
\toprule Benchmark case & Std-PINN & GCR-PINN & LDA-PINN & ACR-PINN\\\midrule
Burgers & $(3.18\pm1.35)\times10^{-3}$ & $(2.86\pm1.15)\times10^{-3}$ & $(7.36\pm4.98)\times10^{-4}$ & $\bm{(6.31\pm4.63)\times10^{-4}}$ \\
Helmholtz (1,4) & $(9.18\pm3.12)\times10^{-2}$ & $(6.65\pm2.11)\times10^{-3}$ & $(1.74\pm0.52)\times10^{-2}$ & $\bm{(2.98\pm1.27)\times10^{-3}}$ \\
Helmholtz (4,4) & $(9.28\pm1.38)\times10^{-2}$ & $(1.49\pm0.51)\times10^{-2}$ & $(6.00\pm0.59)\times10^{-2}$ & $\bm{(9.29\pm2.26)\times10^{-3}}$ \\
Klein--Gordon & $(2.84\pm0.93)\times10^{-2}$ & $(5.09\pm1.51)\times10^{-3}$ & $(8.31\pm2.71)\times10^{-3}$ & $\bm{(1.42\pm0.61)\times10^{-3}}$ \\
Lid-driven cavity & $(4.49\pm0.80)\times10^{-2}$ & $(3.47\pm0.72)\times10^{-2}$ & $(2.72\pm0.76)\times10^{-2}$ & $\bm{(1.96\pm0.08)\times10^{-2}}$ \\
5D Poisson & $(8.32\pm3.33)\times10^{-4}$ & $(5.01\pm0.70)\times10^{-4}$ & $(3.39\pm1.08)\times10^{-4}$ & $\bm{(1.75\pm0.34)\times10^{-4}}$ \\
Nonlinear Schr\"odinger & $(5.99\pm1.65)\times10^{-2}$ & $(2.82\pm0.35)\times10^{-2}$ & $(7.85\pm2.83)\times10^{-3}$ & $\bm{(4.41\pm0.73)\times10^{-3}}$ \\
\bottomrule\end{tabular}}
\end{table}
}

The relative-$L_2$ results in Table~\ref{tab:s17_core_dynamic} show that both individual modifications reduce the mean error relative to Std-PINN in all seven cases, although their relative effectiveness varies by problem. GCR-PINN gives the lower mean among the two single-component methods for the Helmholtz and Klein--Gordon cases, whereas LDA-PINN performs better on Burgers, lid-driven cavity flow, five-dimensional Poisson, and nonlinear Schr\"odinger. Neither component therefore dominates across the full suite.

Combining the two components yields a further reduction in every case. Relative to Std-PINN, ACR-PINN lowers the mean relative $L_2$ error by approximately 56--97\%. It also improves on the stronger single-component method by approximately 14--72\%. The extent of this additional gain varies substantially across benchmarks, showing that the benefit of the combined formulation is problem dependent.

{\small\setlength{\tabcolsep}{3pt}
\begin{table}[H]
\centering\color{black}\captionsetup{font={color=black}}
\caption{Relative $L_\infty$ errors for the seven benchmark cases, reported as mean $\pm$ sample standard deviation over five paired runs. The method with the lowest mean error in each row is highlighted in bold.}
\label{tab:s17_core_linf}
\resizebox{\textwidth}{!}{\begin{tabular}{lcccc}
\toprule Benchmark case & Std-PINN & GCR-PINN & LDA-PINN & ACR-PINN\\\midrule
Burgers & $(1.79\pm0.73)\times10^{-2}$ & $(1.77\pm0.70)\times10^{-2}$ & $(2.61\pm0.93)\times10^{-3}$ & $\bm{(2.14\pm0.79)\times10^{-3}}$ \\
Helmholtz (1,4) & $(3.97\pm0.92)\times10^{-1}$ & $(4.21\pm1.11)\times10^{-2}$ & $(1.21\pm0.34)\times10^{-1}$ & $\bm{(1.71\pm0.34)\times10^{-2}}$ \\
Helmholtz (4,4) & $(5.52\pm1.20)\times10^{-1}$ & $(1.44\pm0.30)\times10^{-1}$ & $(4.81\pm0.55)\times10^{-1}$ & $\bm{(1.27\pm0.68)\times10^{-1}}$ \\
Klein--Gordon & $(4.15\pm1.41)\times10^{-2}$ & $(7.40\pm4.17)\times10^{-3}$ & $(1.06\pm0.33)\times10^{-2}$ & $\bm{(2.49\pm1.83)\times10^{-3}}$ \\
Lid-driven cavity & $(3.89\pm0.43)\times10^{-1}$ & $(2.62\pm0.17)\times10^{-1}$ & $(2.98\pm0.45)\times10^{-1}$ & $\bm{(2.30\pm0.12)\times10^{-1}}$ \\
5D Poisson & $(2.85\pm0.40)\times10^{-3}$ & $(2.67\pm0.43)\times10^{-3}$ & $(2.11\pm0.38)\times10^{-3}$ & $\bm{(1.57\pm0.49)\times10^{-3}}$ \\
Nonlinear Schr\"odinger & $(7.64\pm2.10)\times10^{-2}$ & $(3.35\pm0.40)\times10^{-2}$ & $(1.01\pm0.36)\times10^{-2}$ & $\bm{(5.26\pm0.89)\times10^{-3}}$ \\
\bottomrule\end{tabular}}
\end{table}
}

The relative-$L_\infty$ results in Table~\ref{tab:s17_core_linf} assess whether the domain-wide improvements extend to the largest localized errors. ACR-PINN again gives the lowest observed mean in all seven cases, but its margin over the stronger single-component method is less uniform than for relative $L_2$. The separation is smallest for Helmholtz $(4,4)$ and remains moderate for lid-driven cavity flow, whereas clearer reductions occur for Helmholtz $(1,4)$, Klein--Gordon, five-dimensional Poisson, and nonlinear Schr\"odinger. Thus, a lower aggregate error does not necessarily imply a proportional reduction in the largest local discrepancy. Section~\ref{sec:solution_error_distributions} examines the spatial structure of these errors. Because the results are based on five paired runs, the comparisons are interpreted descriptively in terms of the observed means and run-to-run variability.

{\small\setlength{\tabcolsep}{3pt}
\begin{table}[H]
\centering\color{black}\captionsetup{font={color=black}}
\caption{Mean-squared errors for the seven benchmark cases, reported as mean $\pm$ sample standard deviation over five paired runs. The method with the lowest mean error in each row is highlighted in bold.}
\label{tab:s17_core_mse}
\resizebox{\textwidth}{!}{\begin{tabular}{lcccc}
\toprule Benchmark case & Std-PINN & GCR-PINN & LDA-PINN & ACR-PINN\\\midrule
Burgers & $(4.37\pm3.76)\times10^{-6}$ & $(3.49\pm2.97)\times10^{-6}$ & $(2.79\pm3.70)\times10^{-7}$ & $\bm{(2.15\pm3.24)\times10^{-7}}$ \\
Helmholtz (1,4) & $(2.28\pm1.28)\times10^{-3}$ & $(1.18\pm0.79)\times10^{-5}$ & $(8.08\pm4.66)\times10^{-5}$ & $\bm{(2.52\pm2.34)\times10^{-6}}$ \\
Helmholtz (4,4) & $(2.17\pm0.63)\times10^{-3}$ & $(6.00\pm4.31)\times10^{-5}$ & $(8.98\pm1.77)\times10^{-4}$ & $\bm{(2.24\pm1.13)\times10^{-5}}$ \\
Klein--Gordon & $(1.61\pm0.95)\times10^{-4}$ & $(5.07\pm2.84)\times10^{-6}$ & $(1.37\pm0.81)\times10^{-5}$ & $\bm{(4.25\pm3.43)\times10^{-7}}$ \\
Lid-driven cavity & $(1.45\pm0.50)\times10^{-4}$ & $(8.73\pm3.89)\times10^{-5}$ & $(5.49\pm3.09)\times10^{-5}$ & $\bm{(2.68\pm0.22)\times10^{-5}}$ \\
5D Poisson & $(8.28\pm6.59)\times10^{-6}$ & $(2.70\pm0.79)\times10^{-6}$ & $(1.32\pm0.88)\times10^{-6}$ & $\bm{(3.34\pm1.23)\times10^{-7}}$ \\
Nonlinear Schr\"odinger & $(3.04\pm1.63)\times10^{-3}$ & $(6.46\pm1.55)\times10^{-4}$ & $(5.45\pm4.08)\times10^{-5}$ & $\bm{(1.59\pm0.55)\times10^{-5}}$ \\
\bottomrule\end{tabular}}
\end{table}
}

Table~\ref{tab:s17_core_mse} shows that ACR-PINN attains the lowest mean MSE in all seven cases, with reductions of approximately 82--99.9\% relative to Std-PINN. The largest relative gains occur for Klein--Gordon, nonlinear Schr\"odinger, and the two Helmholtz problems, whereas the improvement is more moderate for lid-driven cavity flow. Compared with the stronger single-component method in each case, ACR-PINN further reduces the mean MSE by approximately 23--92\%. The additional benefit of combining LDA and GCR-PINN is particularly pronounced for Klein--Gordon and Helmholtz $(1,4)$, while remaining clearly problem dependent across the suite.
}

\subsubsection{\textcolor{black}{Solution and Error Distributions}}
\label{sec:solution_error_distributions}

\begin{figure}[H]
\centering
\captionsetup{font={color=black}}
\includegraphics[width=\textwidth]{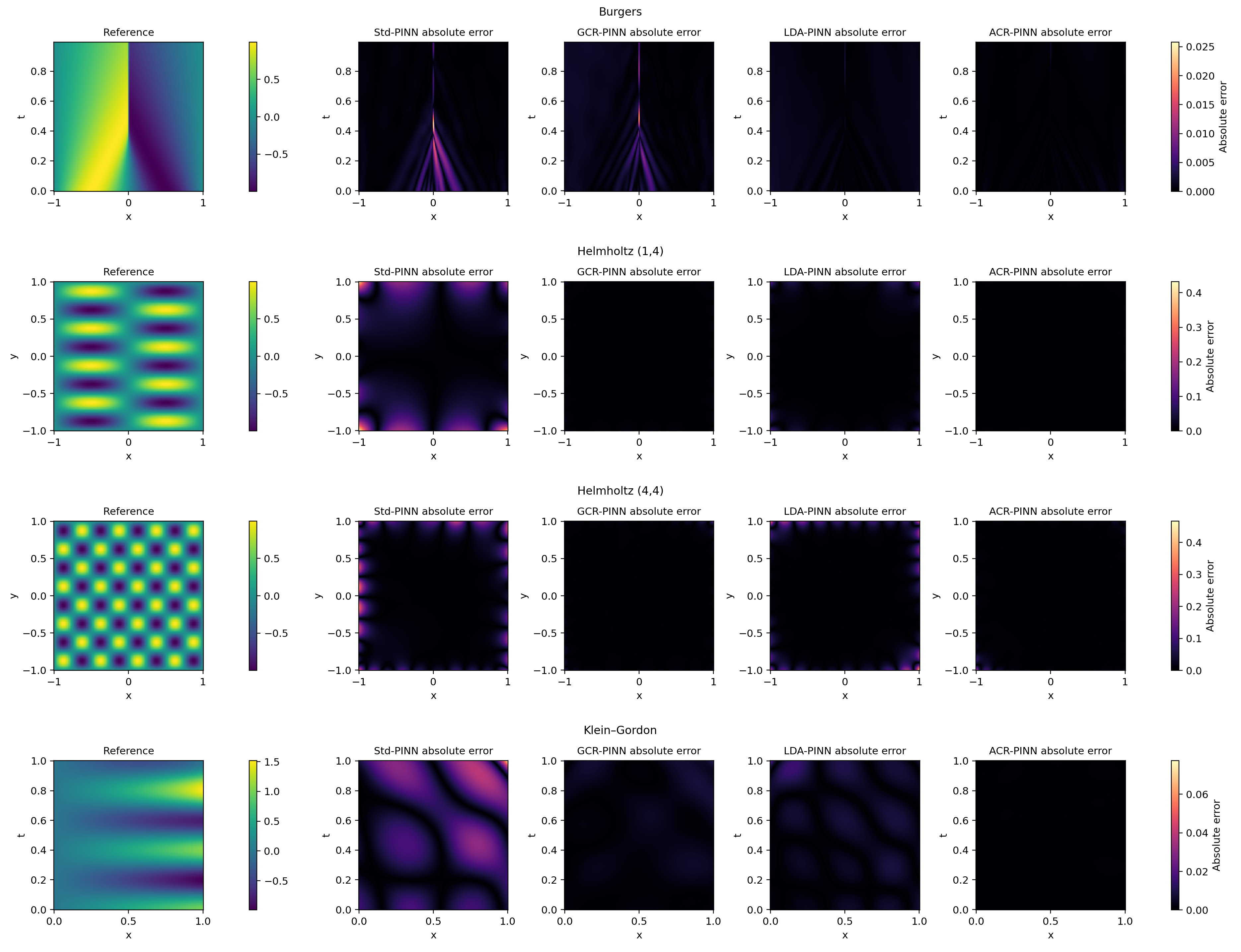}
\caption{
Reference solutions and absolute-error maps for Burgers, Helmholtz $(1,4)$, Helmholtz $(4,4)$, and Klein--Gordon, shown from top to bottom. Columns correspond to the reference solution, Std-PINN, GCR-PINN, LDA-PINN, and ACR-PINN. The four error maps within each row share the same linear color scale. Results are shown for the paired realization with master seed 19018.
}
\label{fig:core_fields_group_a}
\end{figure}

\begin{figure}[H]
\centering
\captionsetup{font={color=black}}
\includegraphics[width=\textwidth]{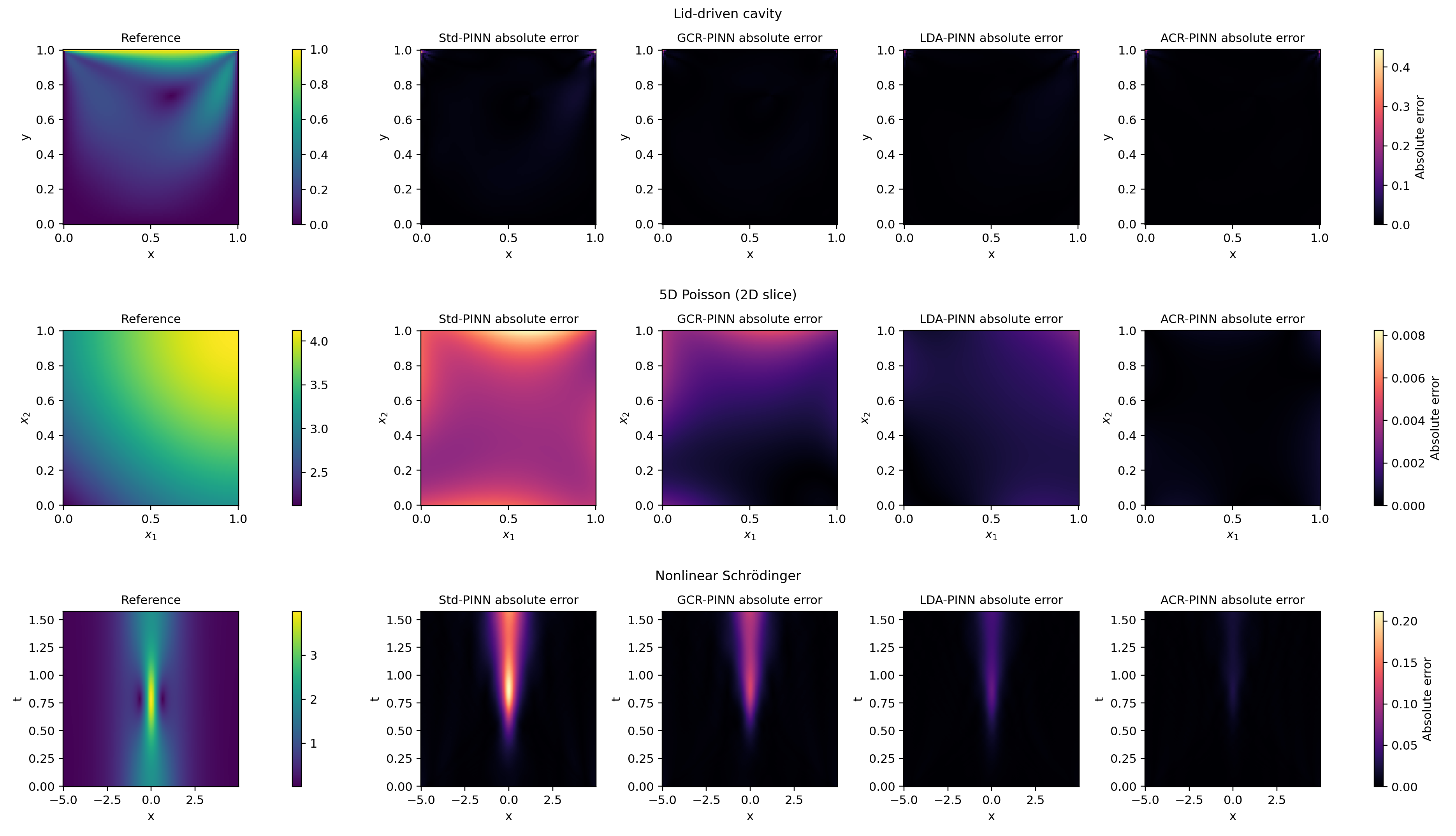}
\caption{
Reference fields and absolute-error maps for lid-driven cavity flow, five-dimensional Poisson, and nonlinear Schr\"odinger, shown from top to bottom. Columns correspond to the reference field, Std-PINN, GCR-PINN, LDA-PINN, and ACR-PINN. The rows show the cavity velocity magnitude, the Poisson slice $x_3=x_4=x_5=0.5$, and the Schr\"odinger reference magnitude $|u|$ with complex-solution error $|u_\theta-u|$, respectively. The four error maps within each row share the same linear color scale. Results are shown for the paired realization with master seed 19018.
}
\label{fig:core_fields_group_b}
\end{figure}

{\color{black}
The aggregate metrics quantify solution accuracy but do not show how the remaining errors are distributed over the computational domain. Figures~\ref{fig:core_fields_group_a} and~\ref{fig:core_fields_group_b} therefore present the reference fields and corresponding absolute-error distributions for one paired run. Figure~\ref{fig:core_fields_group_a} contains the four problems represented directly as real-valued scalar fields over two input coordinates. Figure~\ref{fig:core_fields_group_b} contains the lid-driven cavity, five-dimensional Poisson, and nonlinear Schr\"odinger cases, represented respectively by a derived flow quantity, a two-dimensional slice of the high-dimensional solution, and a complex-valued field. Within each benchmark, the four error maps share the same linear color scale, enabling direct comparison across methods, whereas the scales differ between benchmarks.
}

\FloatBarrier

{\color{black}
The spatial error patterns in Fig.~\ref{fig:core_fields_group_a} complement the aggregate metrics reported above. For Burgers, the dominant discrepancy is concentrated around the steep solution front. LDA-PINN produces a substantially weaker error pattern than GCR-PINN, in agreement with its lower relative $L_2$, relative $L_\infty$, and MSE values, while ACR-PINN further reduces the narrow high-error region. Klein--Gordon exhibits a broader oscillatory distribution. In this case, GCR-PINN provides the stronger single-component result, and the ACR-PINN error field is reduced further over most of the domain.

The Helmholtz cases show a different spatial balance between the two components. For Helmholtz $(1,4)$, GCR-PINN substantially reduces the boundary- and corner-concentrated errors observed for Std-PINN and yields a lower error field than LDA-PINN. ACR-PINN further suppresses these localized structures and reduces both the domain-wide and peak errors relative to GCR-PINN. For Helmholtz $(4,4)$, the errors under Std-PINN and LDA-PINN are distributed more densely along the boundary. GCR-PINN suppresses much of this structure, while ACR-PINN further reduces the overall magnitude. Their comparatively close relative-$L_\infty$ values indicate, however, that the largest residual errors remain of similar scale.

Figure~\ref{fig:core_fields_group_b} shows the corresponding behavior for the remaining benchmarks. In the lid-driven cavity case, the largest errors remain localized near the moving lid and upper corners. ACR-PINN reduces the error over these regions relative to both single-component configurations. For five-dimensional Poisson, the displayed slice shows a broad error distribution for Std-PINN that is reduced by both individual components and further diminished by ACR-PINN; the lower mean relative $L_\infty$ error is consistent with a reduction in the largest local discrepancies. For nonlinear Schr\"odinger, the error is concentrated around the localized wave structure. ACR-PINN produces a narrower and lower-magnitude error region than LDA-PINN, in line with the approximately twofold reduction in the two relative-error measures.

Taken together, the error fields locate the dominant approximation errors and help explain why improvements in domain-wide and maximum local accuracy need not be proportional. The corresponding quantitative comparison is provided by the five-run results in Tables~\ref{tab:s17_core_dynamic}--\ref{tab:s17_core_mse}.
}

\FloatBarrier

\subsection{\textcolor{black}{Ablation and Mechanism Analysis}}
\label{sec:ablation_mechanism}

\subsubsection{\textcolor{black}{Architectural Ablation of LDA}}
\label{sec:lda_ablation}

{\color{black}
We examine whether the observed LDA gain can be explained by model capacity alone and how its performance changes when individual elements of the coordinate pathway are simplified. Burgers, Helmholtz $(4,4)$, and Klein--Gordon are selected to represent nonlinear transport, oscillatory elliptic behavior, and nonlinear wave dynamics, respectively. All architectures are trained with the same summed-loss optimization. The parameter-matched MLP controls for total model size, while four reduced LDA variants probe the roles of per-layer coordinate encoding, dual-branch representation, and input-conditioned feature-wise gating. Table~\ref{tab:lda_controls} summarizes the architectural controls, and Table~\ref{tab:lda_ablation_l2} reports the corresponding relative $L_2$ errors.
}

\begin{table}[H]
\color{black}
\captionsetup{font={color=black}}
\centering
\caption{Architectural controls used in the LDA ablation. All models contain four hidden layers and are trained with the same summed-loss optimization. The parameter-matched MLP has width 140, whereas the remaining variants have width 50.}
\label{tab:lda_controls}
\small
\setlength{\tabcolsep}{6pt}
\resizebox{\textwidth}{!}{
\begin{tabular}{@{}lllr@{}}
\toprule
Variant & Per-layer encoding & Fusion rule & Parameters \\
\midrule
Parameter-matched MLP & None & None & 59,781 \\
Direct-coordinate & Linear projection & Additive injection & 8,251 \\
Single-encoder residual & One nonlinear encoder & Scaled residual & 8,455 \\
Fixed-average & Two nonlinear encoders & Equal-weight residual & 9,055 \\
Static-gate & Two nonlinear encoders & Input-independent feature-wise residual & 9,455 \\
Full LDA & Two nonlinear encoders & Input-conditioned feature-wise residual & 59,655 \\
\bottomrule
\end{tabular}
}
\end{table}

\begin{table}[H]
\color{black}
\captionsetup{font={color=black}}
\centering
\caption{Relative $L_2$ errors of full LDA and its architectural controls. Values are means $\pm$ sample standard deviations over five paired runs; the lowest mean for each benchmark is shown in bold.}
\label{tab:lda_ablation_l2}
\small
\renewcommand{\arraystretch}{1.05}
\resizebox{\textwidth}{!}{%
\begin{tabular}{@{}lccc@{}}
\toprule
Variant & Burgers & Helmholtz $(4,4)$ & Klein--Gordon \\
\midrule
Parameter-matched MLP & $(1.776\pm1.048)\times10^{-3}$ & $(7.060\pm0.288)\times10^{-2}$ & $(9.95\pm2.00)\times10^{-3}$ \\
Direct-coordinate & $(2.889\pm1.186)\times10^{-3}$ & $(1.0383\pm0.3735)\times10^{-1}$ & $(2.583\pm1.030)\times10^{-2}$ \\
Single-encoder residual & $(3.219\pm1.078)\times10^{-3}$ & $(9.231\pm2.049)\times10^{-2}$ & $(2.287\pm1.124)\times10^{-2}$ \\
Fixed-average & $(2.591\pm0.861)\times10^{-3}$ & $(8.449\pm0.887)\times10^{-2}$ & $(3.577\pm0.784)\times10^{-2}$ \\
Static-gate & $(2.054\pm0.133)\times10^{-3}$ & $(7.715\pm1.024)\times10^{-2}$ & $(3.541\pm0.878)\times10^{-2}$ \\
Full LDA & $\bm{(7.36\pm4.98)\times10^{-4}}$ & $\bm{(6.001\pm0.590)\times10^{-2}}$ & $\bm{(8.31\pm2.71)\times10^{-3}}$ \\
\bottomrule
\end{tabular}}
\end{table}

{\color{black}
Full LDA achieves the lowest mean relative $L_2$ error on all three problems. Compared with the parameter-matched MLP, it reduces the mean error by approximately $59\%$, $15\%$, and $16\%$ on Burgers, Helmholtz $(4,4)$, and Klein--Gordon, respectively. The two models contain 59,655 and 59,781 trainable parameters, differing by only about $0.2\%$. The improvement of full LDA over the parameter-matched MLP therefore cannot be explained by parameter count alone, indicating that the way this capacity is organized within the architecture is important.

The reduced variants further show how performance changes as the coordinate pathway is simplified. Direct-coordinate and single-encoder residual remove either the dual representation or its adaptive fusion, while fixed-average and static-gate retain two encoders but restrict how their outputs are combined. None of these reduced configurations matches the performance of full LDA across the three problems. Among them, static-gate gives the lowest mean error on Burgers and Helmholtz $(4,4)$, whereas single-encoder residual performs best on Klein--Gordon. The preferred reduced design therefore varies with the benchmark, while full LDA remains the best-performing architecture in all three cases.

These comparisons support two complementary observations. The parameter-matched control shows that the gain of full LDA is not a consequence of increased parameter count alone. The reduced variants additionally show that simplifying the coordinate pathway consistently degrades the observed mean performance. Taken together, the results support the contribution of the complete layer-specific coordinate-encoding and input-conditioned fusion pathway. The gate analysis in the next subsection examines whether the learned fusion remains functionally active after training.
}

\subsubsection{\textcolor{black}{Learned Gate Behavior and Functional Dependence}}
\label{sec:gate_behavior}

{\color{black}
The architectural ablation shows that the complete LDA pathway attains the lowest mean errors among the tested controls. We next examine whether its input-conditioned gates develop nonuniform branch weights and whether the fitted predictions depend on the learned gating pattern. Burgers, Helmholtz $(4,4)$, and Klein--Gordon are used to compare this behavior across transport, elliptic, and wave problems. For layer $\ell$, feature channel $k$, and input $x$, the normalized binary entropy is defined as
\begin{equation}
\mathcal{H}_{\ell,k}(x)
=
-\frac{1}{\log 2}
\sum_{i=1}^{2}
\alpha_{\ell,k}^{(i)}(x)
\log\alpha_{\ell,k}^{(i)}(x).
\end{equation}
The reported statistic is averaged over evaluation inputs and feature channels and then over the four hidden layers within each run. A value of one corresponds to equal weighting of the two branches, whereas lower values indicate increasingly nonuniform branch allocation.

Functional dependence is examined by keeping all non-gate parameters fixed and perturbing only the learned gate weights. Four interventions are considered: uniform branch weights, input-averaged gate weights for each layer and feature channel, gate values permuted across inputs, and swapped branch weights. Let
$\mathcal{M}=\{\mathrm{unif},\mathrm{mean},\mathrm{perm},\mathrm{swap}\}$
index these interventions, and let
$\varepsilon_{2,s}^{(m)}$
denote the relative $L_2$ error for paired run $s$ under intervention $m$. The intervention ratio is defined as
\begin{equation}
R_{s,m}
=
\frac{\varepsilon_{2,s}^{(m)}}
{\varepsilon_{2,s}^{(\mathrm{learned})}},
\qquad
m\in\mathcal{M},
\label{eq:gate_intervention_ratio}
\end{equation}
where the denominator is evaluated using the learned gates from the same run. Thus, $R_{s,m}>1$ indicates degradation relative to the unmodified model. Since these perturbations are applied after training while the remaining parameters are unchanged, they measure the sensitivity of the fitted prediction to the learned gating pattern rather than the performance of independently retrained alternatives.
}

\begin{figure}[htbp]
\color{black}
\captionsetup{font={color=black}}
\centering
\includegraphics[width=\textwidth]{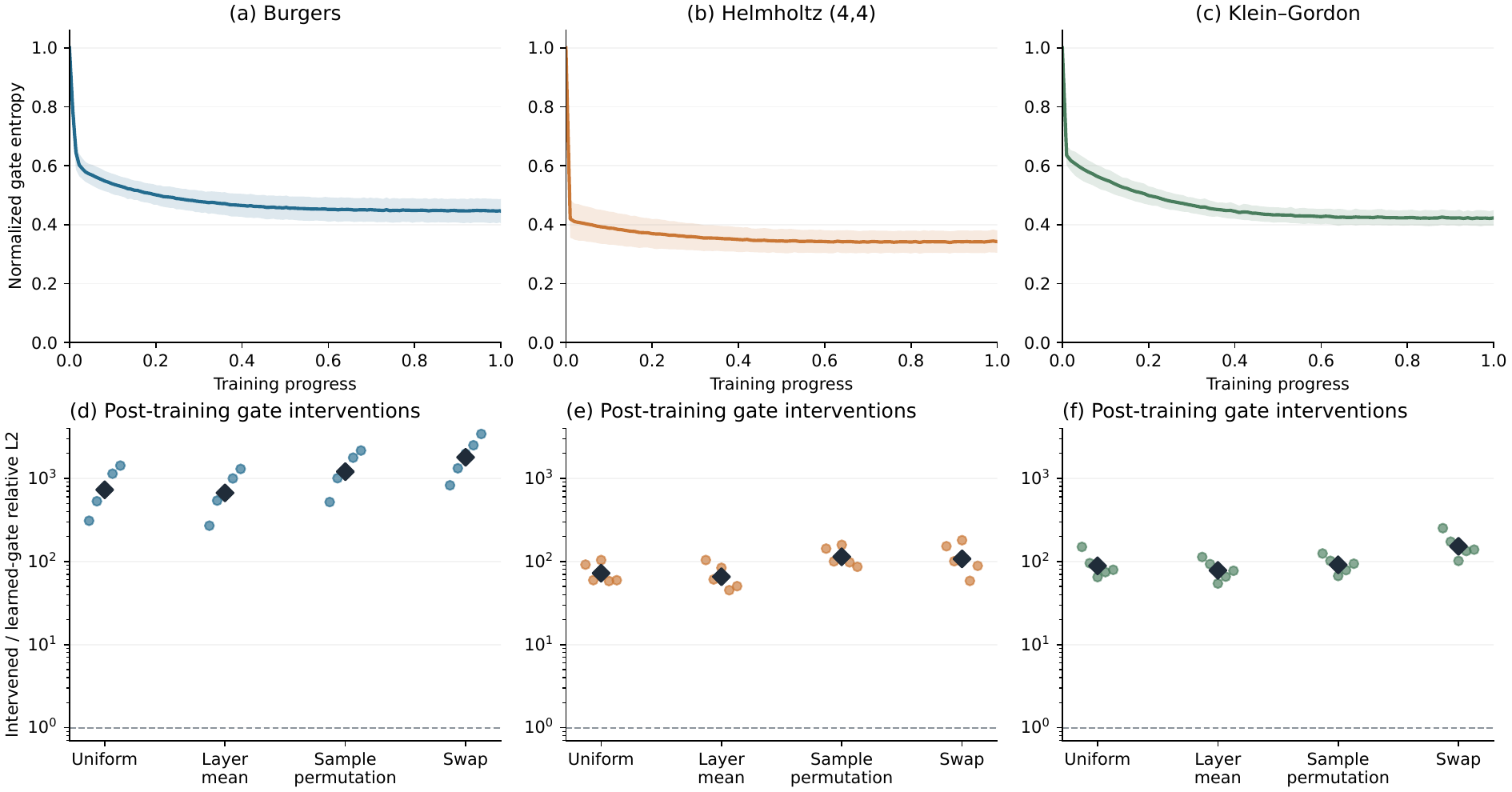}
\caption{Evolution of normalized gate entropy and sensitivity to post-training gate perturbations for Burgers, Helmholtz $(4,4)$, and Klein--Gordon under summed-loss optimization. The upper panels show five-run mean entropy trajectories, with bands indicating one sample standard deviation. In the lower panels, dots denote individual intervention ratios $R_{s,m}$, diamonds denote geometric means, and the dashed line marks the unmodified model at $R_{s,m}=1$; values above one indicate degradation.}
\label{fig:lda_gate_analysis}
\end{figure}

{\color{black}
The upper panels of Fig.~\ref{fig:lda_gate_analysis} show a rapid decrease in entropy early in training followed by gradual stabilization. The final mean entropy is approximately $0.45$ for Burgers, $0.34$ for Helmholtz $(4,4)$, and $0.42$ for Klein--Gordon, well below the equal-weight value of one. The similar evolution across the three problems shows that the branch weights become differentiated early in training and remain nonuniform thereafter. The lower entropy observed for Helmholtz $(4,4)$ indicates a more concentrated allocation between the two branches.

The lower panels provide a direct test of functional dependence. Every intervention increases the relative $L_2$ error in all five runs. The degradation reaches roughly two to three orders of magnitude for Burgers and about one to two orders of magnitude for Helmholtz $(4,4)$ and Klein--Gordon. Replacing the learned gates with uniform or input-averaged weights shows that fixed branch allocations do not preserve the fitted accuracy. The permutation intervention is particularly informative because it retains the learned gate values while breaking their association with the inputs that produced them; the resulting degradation across all three problems indicates that this input dependence is functionally relevant. Swapping the two branch weights also degrades the predictions, showing that the learned allocation depends on the correspondence between the gate weights and the two coordinate representations.

Together with the independently trained controls in Section~\ref{sec:lda_ablation}, these results clarify the role of the LDA gate. The parameter-matched comparison shows that the improvement cannot be explained by model size alone, while the entropy and intervention results show that the fitted models make functional use of nonuniform, input-conditioned feature-wise fusion. The entropy trajectories characterize how the gate departs from uniform weighting during training, and the post-training perturbations show that preserving the learned input--branch correspondence is important for the resulting prediction.
}

\FloatBarrier

\subsubsection{\textcolor{black}{Gradient Geometry and Component Interaction}}
\label{sec:gradient_geometry}
{\color{black}
The preceding ablations show that the complete LDA pathway is associated with improved predictive accuracy. We now examine whether its representation also changes the task-gradient geometry on which GCR-PINN operates. Burgers contains PDE, boundary, and initial-condition tasks, whereas Helmholtz $(4,4)$ couples an oscillatory PDE residual with boundary constraints. Comparing the four core conditions on these two problems allows changes associated with the backbone to be distinguished from those introduced by gradient projection.

For task gradients $g_i=\nabla_\theta\mathcal{L}_i$, let $K$ denote the number of tasks and $Q=K(K-1)/2$ the number of distinct pairs. We characterize their directions and relative magnitudes by
\begin{subequations}
\begin{align}
c_{ij}
&=\frac{g_i^\top g_j}{\|g_i\|_2\|g_j\|_2},\\
\bar c
&=\frac{1}{Q}\sum_{i<j}c_{ij},\\
C
&=\frac{1}{Q}\sum_{i<j}\mathbf{1}\{c_{ij}<0\},\\
S
&=\log_{10}\frac{\max_i\|g_i\|_2}{\min_i\|g_i\|_2}.
\end{align}
\end{subequations}
Here $C$ is the proportion of conflicting task pairs, $\bar c$ measures their average directional alignment, and $S$ summarizes gradient-norm imbalance. All three quantities are evaluated before projection. For GCR-PINN, we additionally define
\begin{equation}
P=\frac{\|g_{\mathrm{CR}}-\sum_i g_i\|_2}{\|\sum_i g_i\|_2},
\label{eq:projection_diagnostic}
\end{equation}
where $g_{\mathrm{CR}}$ is the aggregated GCR-PINN gradient. The ratio $P$ measures the relative change that conflict resolution introduces to the directly summed update, including changes in both direction and magnitude.

Figure~\ref{fig:gradient_geometry_main} traces $C$, $\bar c$, and $P$ over normalized training progress, while Table~\ref{tab:gradient_late} summarizes all four diagnostics over the final third of training. The trajectories show how the gradient geometry evolves, while the late-stage averages characterize its behavior near the end of optimization.

\begin{figure}[H]
\centering\color{black}\captionsetup{font={color=black}}
\includegraphics[width=\textwidth]{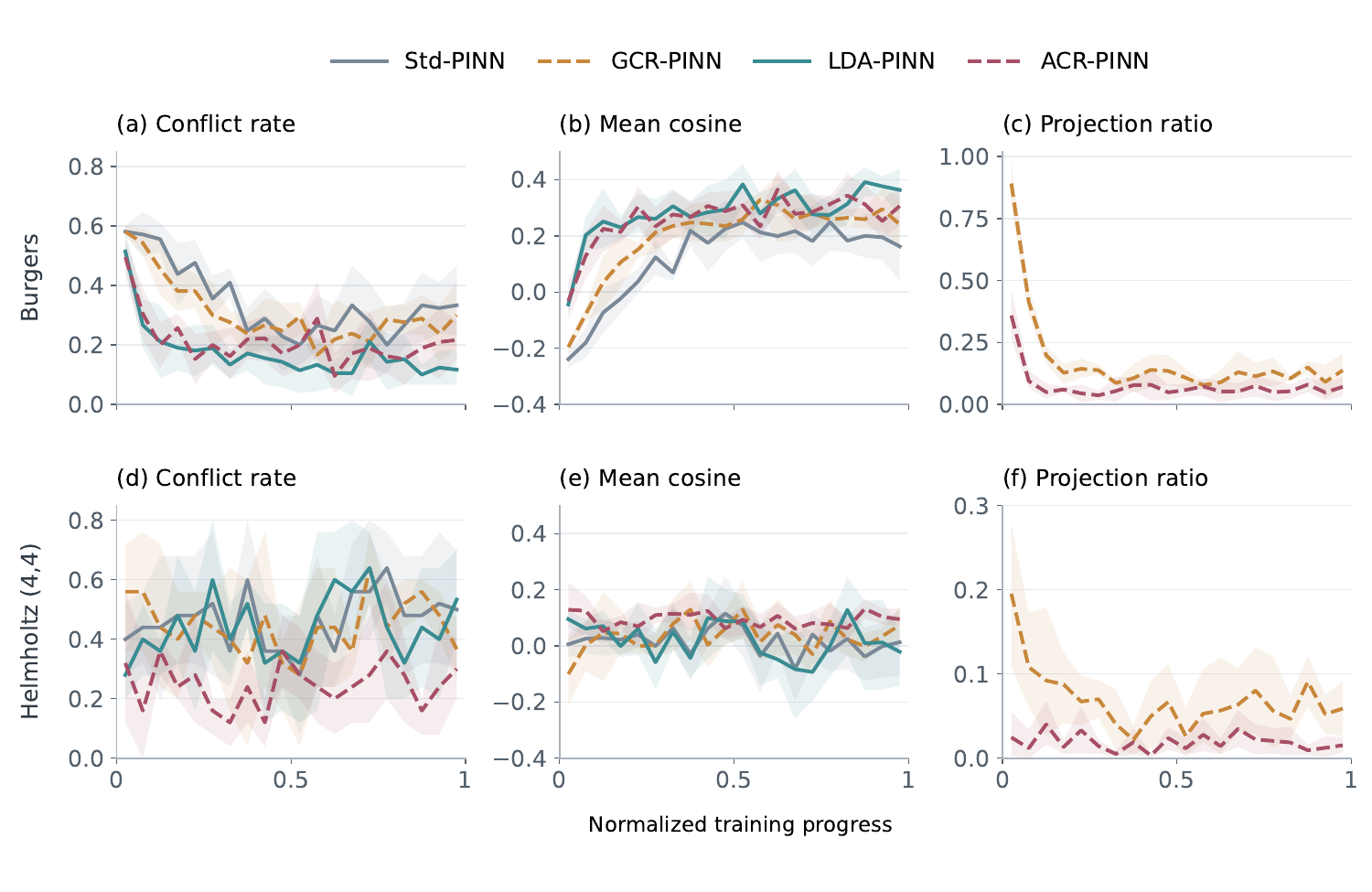}
\caption{Evolution of task-gradient geometry for Burgers (top) and Helmholtz $(4,4)$ (bottom). Columns show the conflict rate, mean pairwise cosine, and projection ratio; curves denote five-run means with pointwise 95\% bootstrap intervals. The projection ratio applies only to GCR-PINN and ACR-PINN. Across the evaluated settings, the LDA backbone is associated with lower conflict rates, while ACR-PINN requires less projection correction than GCR-PINN.}
\label{fig:gradient_geometry_main}
\end{figure}

\begin{table}[H]
\centering\color{black}\captionsetup{font={color=black}}
\caption{Late-stage task-gradient diagnostics for Burgers and Helmholtz $(4,4)$. Each entry reports the mean $\pm$ sample standard deviation across five paired runs, with each diagnostic first averaged over the final third of training within a run. The quantities $C$, $\bar c$, $S$, and $P$ denote conflict rate, mean pairwise cosine, gradient-norm spread, and projection ratio, respectively; a dash indicates that no gradient projection is applied.}
\label{tab:gradient_late}
\small\setlength{\tabcolsep}{4pt}
\resizebox{\textwidth}{!}{\begin{tabular}{@{}llcccc@{}}
\toprule PDE & Model & $C$ & $\bar c$ & $S$ & $P$\\\midrule
Burgers & Std-PINN & $0.297\pm0.088$ & $0.195\pm0.076$ & $1.403\pm0.024$ & -- \\
 & GCR-PINN & $0.261\pm0.050$ & $0.266\pm0.036$ & $1.368\pm0.053$ & $0.120\pm0.044$ \\
 & LDA-PINN & $0.141\pm0.066$ & $0.332\pm0.066$ & $1.292\pm0.066$ & -- \\
 & ACR-PINN & $0.183\pm0.072$ & $0.302\pm0.056$ & $1.258\pm0.097$ & $0.059\pm0.027$ \\
Helmholtz $(4,4)$ & Std-PINN & $0.535\pm0.073$ & $-0.009\pm0.045$ & $2.520\pm0.067$ & -- \\
 & GCR-PINN & $0.482\pm0.085$ & $0.032\pm0.041$ & $3.134\pm0.125$ & $0.062\pm0.015$ \\
 & LDA-PINN & $0.482\pm0.182$ & $-0.008\pm0.130$ & $2.650\pm0.028$ & -- \\
 & ACR-PINN & $0.265\pm0.059$ & $0.092\pm0.023$ & $3.117\pm0.217$ & $0.017\pm0.007$ \\
\bottomrule\end{tabular}}
\end{table}

For Burgers, the upper row of Fig.~\ref{fig:gradient_geometry_main} shows that the conflict rate falls rapidly and the mean cosine becomes positive for all four methods, while the LDA-based curves settle at lower conflict rates and higher mean cosine values. Table~\ref{tab:gradient_late} quantifies this difference. Replacing the MLP with LDA reduces the late-stage conflict rate from $0.297$ to $0.141$ under summed losses and from $0.261$ to $0.183$ under GCR-PINN; it also increases the corresponding mean cosine and reduces the gradient-norm spread $S$. Within the projected models, the projection ratio decreases from $0.120$ to $0.059$. Thus, for Burgers, the LDA backbone produces more compatible gradient directions and a smaller norm spread, while requiring less correction under GCR-PINN.

Helmholtz $(4,4)$ exhibits a different interaction. Under summed losses, changing the backbone produces only a modest reduction in conflict rate, from $0.535$ to $0.482$, with little change in mean cosine. Under projected optimization, however, ACR-PINN reduces the conflict rate from $0.482$ to $0.265$, increases the mean cosine from $0.032$ to $0.092$, and lowers the projection ratio from $0.062$ to $0.017$ relative to GCR-PINN. These differences persist through much of the lower-row trajectories. GCR-PINN and ACR-PINN nevertheless exhibit similar late-stage task-gradient norm spreads ($3.134$ and $3.117$, respectively). This comparison indicates that the improvement is primarily directional: LDA produces a more compatible task-gradient configuration and therefore reduces the amount of correction required by GCR-PINN, rather than merely equalizing gradient magnitudes. Together with the Burgers results, this comparison supports the intended complementarity of the two components while showing that its expression depends on the governing problem and update rule. In Helmholtz $(4,4)$, this complementarity is most clearly reflected in the gradient geometry when LDA and GCR-PINN are combined. The task-resolved curves in~\ref{app:gradient_diagnostics} identify the physical-loss pairs that drive these aggregate trends.

The gradient diagnostics characterize the optimization geometry but do not directly quantify how the two components combine in prediction error. For paired run $s$ and error metric $E$, we therefore define the log-scale interaction contrast
\begin{equation}
I_{s,E}=\log\!\left(
\frac{E_{\mathrm{ACR},s}E_{\mathrm{Std},s}}
{E_{\mathrm{LDA},s}E_{\mathrm{GCR},s}}
\right).
\label{eq:architecture_optimizer_interaction}
\end{equation}
The reference $I_{s,E}=0$ corresponds to multiplicative composition of the two component effects, whereas $I_{s,E}<0$ denotes an additional error reduction.

\begin{table}[H]
\centering\color{black}\captionsetup{font={color=black}}
\caption{Architecture--optimization interaction in prediction error for Burgers and Helmholtz $(4,4)$. Entries report mean paired log contrasts with 95\% bootstrap intervals across five runs. A negative contrast corresponds to an error reduction beyond the multiplicative composition of the individual LDA and GCR-PINN effects.}
\label{tab:gradient_interaction}
\small\begin{tabular}{@{}llcc@{}}
\toprule PDE & Error metric & Mean interaction & 95\% bootstrap interval\\\midrule
Burgers & Relative $L_2$ & $-0.057$ & $[-0.630, 0.517]$ \\
 & Relative $L_\infty$ & $-0.206$ & $[-0.681, 0.400]$ \\
Helmholtz $(4,4)$ & Relative $L_2$ & $-0.019$ & $[-0.507, 0.444]$ \\
 & Relative $L_\infty$ & $-0.085$ & $[-0.240, 0.119]$ \\
\bottomrule\end{tabular}
\end{table}

All four mean interaction estimates in Table~\ref{tab:gradient_interaction} are negative. The largest observed contrast occurs for the Burgers relative $L_\infty$ error, whereas the two Helmholtz estimates lie closer to the multiplicative reference. The consistently negative point estimates provide directional evidence of complementary effects between LDA and GCR-PINN. This interpretation agrees with the gradient diagnostics: LDA changes the task-gradient configuration, while GCR-PINN acts on the conflicts that remain. The magnitude of the observed interaction nevertheless varies across problems and error measures.
}

\subsection{\textcolor{black}{Comparison with Representative Methods}}
\label{sec:method_comparisons}
{\color{black}
Two complementary comparisons are used to assess the proposed ACR-PINN framework relative to existing PINN methods. The first isolates the optimization component by comparing GCR-PINN with representative gradient-handling strategies on a common MLP backbone. The second evaluates the complete ACR-PINN against representative PINN baselines under a common fixed-sampling protocol. Together, these comparisons distinguish the contribution of conflict-resolved optimization from the performance of the integrated framework.
}

\subsubsection{\textcolor{black}{Optimization Strategies on a Common MLP Backbone}}
{\color{black}
To isolate the effect of task-gradient treatment, all optimization strategies are evaluated with the same MLP backbone. Burgers couples PDE, initial-condition, and boundary-condition tasks in a nonlinear time-dependent problem, whereas Helmholtz $(4,4)$ combines an oscillatory interior residual with boundary constraints. The two cases therefore examine whether the optimization strategies remain effective across distinct task structures without conflating their effects with changes in network architecture.

The comparison includes Std-PINN with direct gradient summation, GCR-PINN with conditional pairwise projection~\cite{yu2020gradient}, the multiple-gradient descent algorithm (MGDA) with minimum-norm gradient combination~\cite{sener2018multitask}, GradNorm with gradient-norm balancing~\cite{chen2018gradnorm}, and neural tangent kernel (NTK) trace-ratio weighting~\cite{wang2022and}. All methods follow the paired five-run dynamic-resampling protocol specified in Table~\ref{tab:core_protocol}.
\begin{table}[htbp]
\centering\color{black}\small\captionsetup{font={color=black}}
\caption{Prediction errors of alternative gradient-handling strategies on a common MLP backbone. Entries are means $\pm$ sample standard deviations across five paired runs; the lowest mean for each problem and metric is shown in bold.}
\label{tab:ntk_five_seed}
\setlength{\tabcolsep}{5pt}
\resizebox{\textwidth}{!}{\begin{tabular}{lccc}
\toprule Optimization strategy & Relative $L_2$ & Relative $L_\infty$ & MSE\\
\midrule
\multicolumn{4}{c}{Burgers} \\
\midrule
Std-PINN (direct summation) & $(3.18\pm1.35)\times10^{-3}$ & $(1.79\pm0.73)\times10^{-2}$ & $(4.37\pm3.76)\times10^{-6}$ \\
MGDA & $(2.13\pm0.51)\times10^{-2}$ & $(1.61\pm0.52)\times10^{-1}$ & $(1.79\pm0.73)\times10^{-4}$ \\
GradNorm & $(4.93\pm1.51)\times10^{-3}$ & $(2.95\pm1.24)\times10^{-2}$ & $(9.87\pm5.93)\times10^{-6}$ \\
NTK weighting & $(2.73\pm0.21)\times10^{-1}$ & $(8.23\pm0.47)\times10^{-1}$ & $(2.83\pm0.41)\times10^{-2}$ \\
GCR-PINN & $\mathbf{(2.86\pm1.15)\times10^{-3}}$ & $\mathbf{(1.77\pm0.70)\times10^{-2}}$ & $\mathbf{(3.49\pm2.97)\times10^{-6}}$ \\
\midrule
\multicolumn{4}{c}{Helmholtz $(4,4)$} \\
\midrule
Std-PINN (direct summation) & $(9.28\pm1.38)\times10^{-2}$ & $(5.52\pm1.20)\times10^{-1}$ & $(2.17\pm0.63)\times10^{-3}$ \\
MGDA & $(1.17\pm0.75)\times10^{-1}$ & $\mathbf{(1.29\pm0.66)\times10^{-1}}$ & $(4.49\pm4.42)\times10^{-3}$ \\
GradNorm & $(1.88\pm0.23)\times10^{-1}$ & $(8.10\pm2.59)\times10^{-1}$ & $(8.84\pm2.21)\times10^{-3}$ \\
NTK weighting & $(1.33\pm0.37)\times10^{0}$ & $(1.87\pm0.28)\times10^{0}$ & $(4.62\pm0.21)\times10^{-1}$ \\
GCR-PINN & $\mathbf{(1.49\pm0.51)\times10^{-2}}$ & $(1.44\pm0.30)\times10^{-1}$ & $\mathbf{(6.00\pm4.31)\times10^{-5}}$ \\
\bottomrule\end{tabular}}
\end{table}
Table~\ref{tab:ntk_five_seed} shows that the relative effectiveness of the optimization strategies depends on the task structure. On Burgers, where direct summation already yields comparatively small errors, GCR-PINN nevertheless attains the lowest mean value for all three metrics. Its advantage becomes more pronounced on Helmholtz $(4,4)$: relative $L_2$ decreases from $9.28\times10^{-2}$ for Std-PINN to $1.49\times10^{-2}$, while MSE decreases from $2.17\times10^{-3}$ to $6.00\times10^{-5}$. MGDA gives a slightly lower mean relative $L_\infty$ error on Helmholtz, but its relative $L_2$ error and MSE are markedly higher, indicating less consistent accuracy over the full solution field. GradNorm and NTK weighting also remain above GCR-PINN in both global error measures. Across the six problem--metric combinations, GCR-PINN records five lowest means and one second-lowest mean, including the lowest relative $L_2$ and MSE on both equations. Because the network architecture and training protocol are held fixed, these results show that conditional conflict resolution independently improves the overall effectiveness of multitask gradient updates, thereby validating the optimization design adopted in ACR-PINN.
}

\FloatBarrier

\subsubsection{\textcolor{black}{Evaluation against Representative PINN Baselines}}
\label{sec:strong_baselines}
{\color{black}
Having isolated the optimization component, we next evaluate the complete ACR-PINN against representative PINN baselines on Burgers and Helmholtz $(4,4)$. The comparison spans standard training, the individual architectural and optimization components, alternative input representations, and adaptive residual- or gradient-weighting strategies. Specifically, it includes Std-PINN, GCR-PINN, LDA-PINN, the modified MLP with gradient-statistics learning-rate annealing (M-MLP+LRA)~\cite{wang2021understanding}, a Fourier-feature PINN (FF-PINN)~\cite{tancik2020fourier}, a residual-based-attention PINN (RBA-PINN)~\cite{anagnostopoulos2024rba}, and a balanced-residual-decay-rate PINN (BRDR-PINN)~\cite{chen2025brdr}.

Because the pointwise weights used by RBA-PINN and BRDR-PINN are associated with individual collocation points, all methods are retrained using the same fixed collocation sets. The iteration budget, Adam schedule, evaluation set, and paired-run design are otherwise held constant. The results in this subsection therefore constitute a separate controlled comparison and should not be combined directly with the dynamic-resampling results in Section~\ref{sec:core_seven}. Implementation details are provided in~\ref{app:baseline_configuration}.

\begin{table}[H]
\centering
\color{black}
\captionsetup{font={color=black}}
\caption{Prediction errors of representative PINN methods under a common fixed-sampling protocol. Entries are means $\pm$ sample standard deviations over five paired runs; the smallest mean within each problem--metric combination is shown in bold.}
\label{tab:strong_baseline_results}
\small
\setlength{\tabcolsep}{4pt}
\resizebox{\textwidth}{!}{\begin{tabular}{@{}lccc@{}}
\toprule
Method & Relative $L_2$ & Relative $L_\infty$ & MSE \\
\midrule
\multicolumn{4}{c}{Burgers} \\
\midrule
Std-PINN & $(6.83\pm3.42)\times10^{-3}$ & $(6.27\pm3.41)\times10^{-2}$ & $(2.12\pm1.54)\times10^{-5}$ \\
GCR-PINN & $(3.72\pm1.89)\times10^{-3}$ & $(3.43\pm2.02)\times10^{-2}$ & $(6.30\pm5.78)\times10^{-6}$ \\
LDA-PINN & $(6.17\pm10.42)\times10^{-3}$ & $(6.68\pm12.00)\times10^{-2}$ & $(4.72\pm10.31)\times10^{-5}$ \\
ACR-PINN & $(1.15\pm0.59)\times10^{-3}$ & $\mathbf{(8.81\pm5.58)\times10^{-3}}$ & $(6.05\pm4.81)\times10^{-7}$ \\
M-MLP+LRA & $(7.14\pm0.51)\times10^{-1}$ & $(1.08\pm0.07)\times10^{0}$ & $(1.93\pm0.28)\times10^{-1}$ \\
FF-PINN & $(1.42\pm0.32)\times10^{-1}$ & $(1.36\pm0.20)\times10^{0}$ & $(7.91\pm3.74)\times10^{-3}$ \\
RBA-PINN & $(5.07\pm2.46)\times10^{-3}$ & $(5.30\pm3.44)\times10^{-2}$ & $(1.15\pm0.97)\times10^{-5}$ \\
BRDR-PINN & $\mathbf{(1.10\pm0.31)\times10^{-3}}$ & $(1.01\pm0.46)\times10^{-2}$ & $\mathbf{(4.85\pm2.40)\times10^{-7}}$ \\
\midrule
\multicolumn{4}{c}{Helmholtz $(4,4)$} \\
\midrule
Std-PINN & $(1.18\pm0.18)\times10^{-1}$ & $(7.95\pm0.64)\times10^{-1}$ & $(3.49\pm1.13)\times10^{-3}$ \\
GCR-PINN & $(1.49\pm0.77)\times10^{-2}$ & $(1.82\pm0.94)\times10^{-1}$ & $(6.69\pm6.88)\times10^{-5}$ \\
LDA-PINN & $(6.97\pm1.33)\times10^{-2}$ & $(5.42\pm2.17)\times10^{-1}$ & $(1.24\pm0.51)\times10^{-3}$ \\
ACR-PINN & $\mathbf{(7.91\pm1.50)\times10^{-3}}$ & $\mathbf{(1.11\pm0.70)\times10^{-1}}$ & $\mathbf{(1.59\pm0.55)\times10^{-5}}$ \\
M-MLP+LRA & $(1.001\pm0.001)\times10^{0}$ & $(1.05\pm0.03)\times10^{0}$ & $(2.48\pm0.01)\times10^{-1}$ \\
FF-PINN & $(5.78\pm0.52)\times10^{-2}$ & $(3.62\pm1.28)\times10^{-1}$ & $(8.31\pm1.48)\times10^{-4}$ \\
RBA-PINN & $(1.81\pm0.21)\times10^{-2}$ & $(1.69\pm0.66)\times10^{-1}$ & $(8.21\pm1.87)\times10^{-5}$ \\
BRDR-PINN & $(3.49\pm0.48)\times10^{-2}$ & $(3.24\pm1.04)\times10^{-1}$ & $(3.06\pm0.81)\times10^{-4}$ \\
\bottomrule
\end{tabular}}
\end{table}

Table~\ref{tab:strong_baseline_results} first demonstrates the benefit of combining the two proposed components. On both equations, ACR-PINN reduces all three mean errors relative to either GCR-PINN or LDA-PINN alone. For Burgers, its relative $L_2$ error is $1.149\times10^{-3}$, compared with $3.718\times10^{-3}$ for GCR-PINN and $6.168\times10^{-3}$ for LDA-PINN. The same ordering holds for relative $L_\infty$ and MSE. On Helmholtz $(4,4)$, ACR-PINN decreases the relative $L_2$ error to $7.905\times10^{-3}$, approximately half the GCR-PINN value and nearly one order of magnitude below the LDA-PINN value. These consistent reductions indicate that the architectural and optimization components make complementary contributions under the controlled protocol.

The broader comparison further shows that ACR-PINN remains among the leading methods across both equations. On Burgers, ACR-PINN and BRDR-PINN attain comparable global accuracy: their relative-$L_2$ means differ by approximately $4.6\%$, and both MSE values are of order $10^{-7}$, more than one order of magnitude below the next-lowest result. ACR-PINN additionally gives the lowest mean relative $L_\infty$ error, reducing it from $1.008\times10^{-2}$ for BRDR-PINN to $8.813\times10^{-3}$. On Helmholtz $(4,4)$, ACR-PINN records the lowest mean for all three metrics; its relative $L_2$ error is lower than those of GCR-PINN and RBA-PINN, and the corresponding relative-$L_\infty$ and MSE values exhibit the same ordering. The advantage on this more oscillatory problem therefore extends across both global and maximum-error measures.

\begin{figure}[H]
\centering
\captionsetup{font={color=black}}
\includegraphics[width=\textwidth]{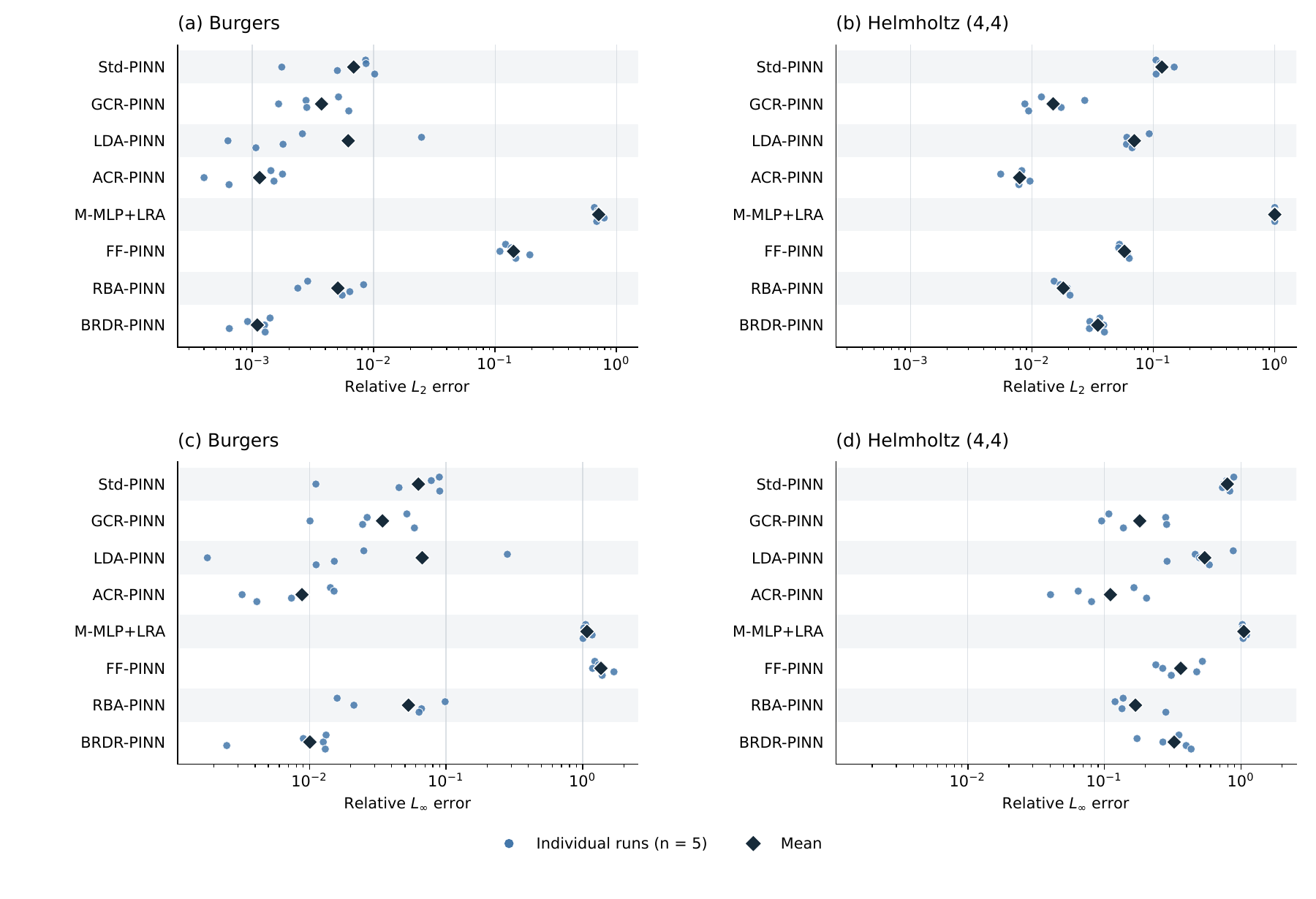}
\caption{Run-wise relative prediction errors under fixed sampling. Columns correspond to Burgers and Helmholtz $(4,4)$, while rows report relative $L_2$ and relative $L_\infty$. Circles denote individual paired runs and diamonds denote arithmetic means; all horizontal axes use logarithmic scales. ACR-PINN remains within the low-error range on both equations.}
\label{fig:strong_baseline_runs}
\end{figure}
\FloatBarrier

Figure~\ref{fig:strong_baseline_runs} complements the aggregate errors in Table~\ref{tab:strong_baseline_results} by showing their run-to-run distributions. On Burgers, the relative-$L_2$ distributions of ACR-PINN and BRDR-PINN overlap substantially, indicating that their small difference in mean global error is modest relative to the variation across runs. ACR-PINN nevertheless remains concentrated in the low-error region and attains the lowest mean relative $L_\infty$ error. The larger standard deviations of LDA-PINN originate from a more dispersed set of Burgers runs. On Helmholtz $(4,4)$, the ACR-PINN results remain toward the low-error end for both metrics, despite partial overlap with GCR-PINN and RBA-PINN in relative $L_\infty$. Taken together, the table and figure show that ACR-PINN achieves leading or near-leading predictive accuracy on both equations and remains consistently competitive in terms of both global and maximum errors. These results further demonstrate the effectiveness of combining architectural enhancement with gradient-conflict resolution.
}

\FloatBarrier

\subsection{\textcolor{black}{Robustness and Extended-Time Assessment}}
\label{sec:robustness_extended_time}
{\color{black}
We next examine whether the relative accuracy of the four core methods persists under changes in sampling, initialization, and temporal horizon. For the time-dependent Klein--Gordon problem, we additionally assess whether explicitly separating the spatial and temporal coordinate encodings improves upon the joint LDA representation.

\paragraph{Sampling and initialization sensitivity}
The oscillatory solution and coupled PDE--boundary constraints of Helmholtz $(4,4)$ provide a demanding setting for assessing sensitivity to collocation sampling and parameter initialization. Table~\ref{tab:sampling_sensitivity} summarizes a crossed design with three seeds for each source of randomness, giving nine matched combinations per method.

\begin{table}[htbp]
\centering
\color{black}
\captionsetup{font={color=black}}
\caption{Helmholtz $(4,4)$ errors across the nine combinations of three sampling seeds and three initialization seeds. Entries are means $\pm$ sample standard deviations over the nine combinations; the lowest mean in each column is shown in bold.}
\label{tab:sampling_sensitivity}
\small
\setlength{\tabcolsep}{5pt}
\resizebox{\textwidth}{!}{%
\begin{tabular}{lccc}
\toprule
Method & Relative $L_2$ & Relative $L_\infty$ & MSE \\
\midrule
Std-PINN & $(9.73\pm2.22)\times10^{-2}$ & $(6.49\pm1.69)\times10^{-1}$ & $(2.45\pm1.09)\times10^{-3}$ \\
GCR-PINN & $(1.61\pm0.57)\times10^{-2}$ & $(1.74\pm0.70)\times10^{-1}$ & $(7.16\pm5.58)\times10^{-5}$ \\
LDA-PINN & $(5.89\pm0.71)\times10^{-2}$ & $(4.29\pm0.59)\times10^{-1}$ & $(8.70\pm1.99)\times10^{-4}$ \\
ACR-PINN & $\mathbf{(9.87\pm3.94)\times10^{-3}}$ & $\mathbf{(8.83\pm6.13)\times10^{-2}}$ & $\mathbf{(2.75\pm2.29)\times10^{-5}}$ \\
\bottomrule
\end{tabular}}
\end{table}

Across the nine matched combinations, ACR-PINN retains the lowest observed mean for all three metrics. Relative to GCR-PINN, the next-best method, it reduces the mean relative $L_2$, relative $L_\infty$, and MSE by $38.8\%$, $49.1\%$, and $61.5\%$, respectively. The relative-$L_2$ variance decomposition assigns $28.4\%$ to sampling, $39.7\%$ to initialization, and $32.0\%$ to their interaction and residual variation, indicating that both sampling and initialization contribute materially to the observed variation.

\paragraph{Extended-time performance}
Klein--Gordon is considered because its nonlinear wave dynamics allow the temporal horizon to be varied without changing the governing structure. Separate models are trained and evaluated on $T=1$, $2$, and $4$; this comparison therefore measures performance over increasingly long domains rather than extrapolation beyond the training interval. Table~\ref{tab:kg_longtime_four_methods} reports the full-domain errors, while Fig.~\ref{fig:kg_longtime_relative_l2} resolves the temporal distributions of relative $L_2$.

\begin{table}[htbp]
\centering
\color{black}
\captionsetup{font={color=black}}
\caption{Klein--Gordon errors for training horizons $T=1$, $2$, and $4$. Entries are means $\pm$ sample standard deviations over three paired runs; the lowest mean for each horizon and metric is shown in bold.}
\label{tab:kg_longtime_four_methods}
\scriptsize
\resizebox{\textwidth}{!}{%
\renewcommand{\arraystretch}{1.05}
\begin{tabular}{@{}c l c c c@{}}
\toprule
$T$ & Method & Relative $L_2$ & Relative $L_\infty$ & MSE \\
\midrule
1 & Std-PINN & $(3.11\pm1.04)\times10^{-2}$ & $(5.03\pm1.06)\times10^{-2}$ & $(1.90\pm1.08)\times10^{-4}$ \\
  & GCR-PINN & $(5.84\pm1.19)\times10^{-3}$ & $(8.78\pm4.99)\times10^{-3}$ & $(6.41\pm2.69)\times10^{-6}$ \\
  & LDA-PINN & $(1.13\pm0.30)\times10^{-2}$ & $(3.69\pm0.53)\times10^{-2}$ & $(2.47\pm1.25)\times10^{-5}$ \\
  & ACR-PINN & $\mathbf{(2.06\pm0.53)\times10^{-3}}$ & $\mathbf{(2.17\pm0.46)\times10^{-3}}$ & $\mathbf{(8.10\pm3.82)\times10^{-7}}$ \\
\midrule
2 & Std-PINN & $(4.79\pm1.52)\times10^{-2}$ & $(3.89\pm0.27)\times10^{-2}$ & $(3.73\pm2.12)\times10^{-3}$ \\
  & GCR-PINN & $(6.73\pm3.33)\times10^{-3}$ & $(4.92\pm1.54)\times10^{-3}$ & $(8.04\pm6.76)\times10^{-5}$ \\
  & LDA-PINN & $(1.87\pm0.93)\times10^{-2}$ & $(1.29\pm0.61)\times10^{-2}$ & $(6.23\pm6.07)\times10^{-4}$ \\
  & ACR-PINN & $\mathbf{(1.86\pm0.33)\times10^{-3}}$ & $\mathbf{(1.24\pm0.63)\times10^{-3}}$ & $\mathbf{(5.41\pm1.83)\times10^{-6}}$ \\
\midrule
4 & Std-PINN & $(2.52\pm0.21)\times10^{-1}$ & $(1.52\pm0.37)\times10^{-1}$ & $(5.47\pm0.90)\times10^{0}$ \\
  & GCR-PINN & $(6.32\pm0.78)\times10^{-2}$ & $(3.38\pm0.29)\times10^{-2}$ & $(3.47\pm0.82)\times10^{-1}$ \\
  & LDA-PINN & $(1.28\pm0.55)\times10^{-1}$ & $(5.79\pm1.28)\times10^{-2}$ & $(1.59\pm1.37)\times10^{0}$ \\
  & ACR-PINN & $\mathbf{(3.87\pm0.83)\times10^{-2}}$ & $\mathbf{(2.25\pm0.40)\times10^{-2}}$ & $\mathbf{(1.33\pm0.57)\times10^{-1}}$ \\
\bottomrule
\end{tabular}}
\end{table}

\begin{figure}[H]
\centering
\captionsetup{labelfont={color=black},textfont={color=black}}
\includegraphics[width=\textwidth]{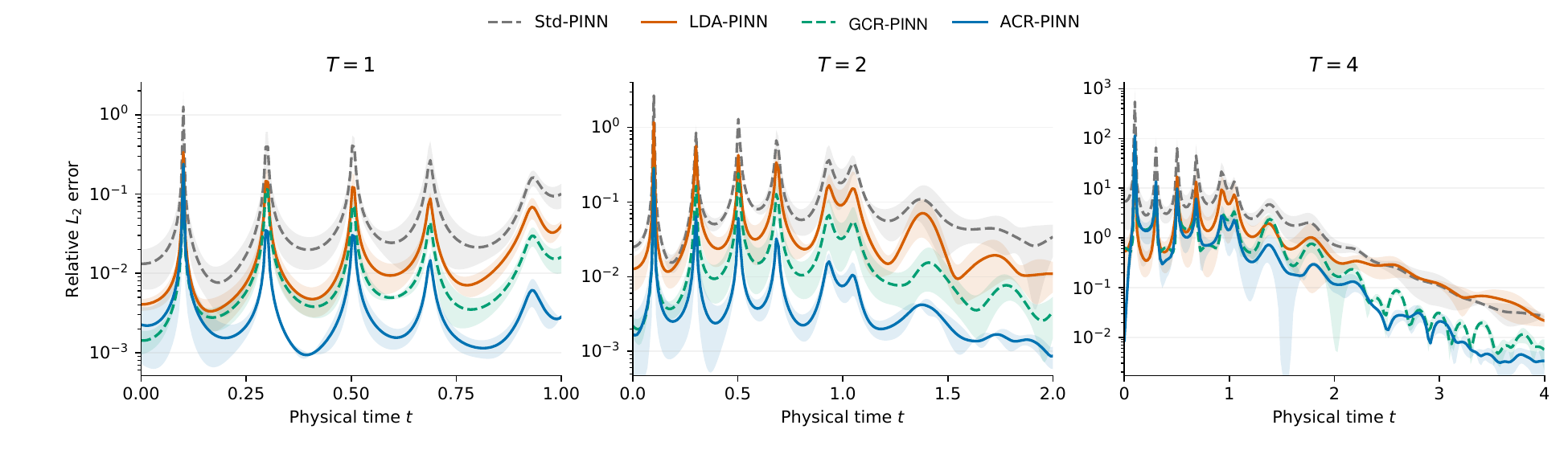}
\caption{Time-resolved relative $L_2$ error for separately trained Klein--Gordon problems with $T=1$, $2$, and $4$. Curves are means over three paired runs, and bands span one sample standard deviation.}
\label{fig:kg_longtime_relative_l2}
\end{figure}

ACR-PINN attains the lowest observed mean error among the four core methods at each horizon and reduces the mean relative $L_2$ error over GCR-PINN by $64.7\%$, $72.3\%$, and $38.7\%$ for $T=1$, $2$, and $4$, respectively. Figure~\ref{fig:kg_longtime_relative_l2} shows that this advantage is distributed over the temporal domain rather than confined to the aggregate metric. At $T=1$ and $T=2$, ACR-PINN remains near the lower error envelope through most of the interval, including the recurrent local peaks. At $T=4$, the between-run spread increases and the early-time separation from GCR-PINN narrows, although ACR-PINN remains below Std-PINN and LDA-PINN over most of the domain and returns to the lowest error range at later times. The full-domain errors nevertheless increase at $T=4$, indicating that ACR-PINN mitigates rather than eliminates error accumulation over longer temporal domains.

\paragraph{Space--time encoding factorization}
The standard LDA encoder acts on the joint coordinate $\boldsymbol q=(x,t)$. To assess whether the distinct roles of space and time warrant explicit factorization, we compare it with a variant that encodes $x$ and $t$ separately before gated fusion. Both variants use four hidden layers of width 50 under the same Klein--Gordon sampling and optimization conditions. The error comparison uses three paired runs for direct-sum and GCR-PINN updates; the joint and separate models contain 59,655 and 59,255 trainable parameters, respectively. A separate five-run analysis at $T=1$ examines how the factorized model allocates its gate weights across depth.

\begin{table}[H]
\centering
\color{black}
\captionsetup{labelfont={color=black},textfont={color=black}}
\scriptsize
\caption{Klein--Gordon errors with joint and separate space--time encoding. Entries are means $\pm$ sample standard deviations over three paired runs after the predefined training budget, evaluated on an independent $200\times200$ grid; the smaller mean within each matched encoding pair is shown in bold.}
\label{tab:kg_joint_separate}
\resizebox{\textwidth}{!}{%
\renewcommand{\arraystretch}{1.05}
\begin{tabular}{@{}c l l c c c@{}}
\toprule
$T$ & Optimization & Encoding & Relative $L_2$ & Relative $L_\infty$ & MSE \\
\midrule
1 & Direct sum & Joint    & $\mathbf{(1.13\pm0.30)\times10^{-2}}$ & $(3.69\pm0.53)\times10^{-2}$ & $\mathbf{(2.47\pm1.25)\times10^{-5}}$ \\
  &            & Separate & $(1.40\pm0.36)\times10^{-2}$ & $\mathbf{(2.99\pm0.89)\times10^{-2}}$ & $(3.74\pm1.97)\times10^{-5}$ \\
  & GCR-PINN   & Joint    & $\mathbf{(2.06\pm0.53)\times10^{-3}}$ & $(2.17\pm0.46)\times10^{-3}$ & $\mathbf{(8.10\pm3.82)\times10^{-7}}$ \\
  &            & Separate & $(2.17\pm0.66)\times10^{-3}$ & $\mathbf{(2.05\pm0.54)\times10^{-3}}$ & $(9.12\pm5.45)\times10^{-7}$ \\
\midrule
2 & Direct sum & Joint    & $\mathbf{(1.87\pm0.93)\times10^{-2}}$ & $\mathbf{(1.29\pm0.61)\times10^{-2}}$ & $\mathbf{(6.23\pm6.07)\times10^{-4}}$ \\
  &            & Separate & $(2.13\pm0.23)\times10^{-2}$ & $(2.16\pm1.68)\times10^{-2}$ & $(6.99\pm1.55)\times10^{-4}$ \\
  & GCR-PINN   & Joint    & $(1.86\pm0.33)\times10^{-3}$ & $(1.24\pm0.63)\times10^{-3}$ & $(5.41\pm1.83)\times10^{-6}$ \\
  &            & Separate & $\mathbf{(1.73\pm0.74)\times10^{-3}}$ & $\mathbf{(1.23\pm0.38)\times10^{-3}}$ & $\mathbf{(5.15\pm3.94)\times10^{-6}}$ \\
\midrule
4 & Direct sum & Joint    & $(1.28\pm0.55)\times10^{-1}$ & $(5.79\pm1.28)\times10^{-2}$ & $(1.59\pm1.37)\times10^{0}$ \\
  &            & Separate & $\mathbf{(1.06\pm0.07)\times10^{-1}}$ & $\mathbf{(5.21\pm0.94)\times10^{-2}}$ & $\mathbf{(9.72\pm1.33)\times10^{-1}}$ \\
  & GCR-PINN   & Joint    & $(3.87\pm0.83)\times10^{-2}$ & $\mathbf{(2.25\pm0.40)\times10^{-2}}$ & $(1.33\pm0.57)\times10^{-1}$ \\
  &            & Separate & $\mathbf{(3.79\pm0.13)\times10^{-2}}$ & $(3.04\pm0.02)\times10^{-2}$ & $\mathbf{(1.24\pm0.09)\times10^{-1}}$ \\
\bottomrule
\end{tabular}}
\end{table}

Table~\ref{tab:kg_joint_separate} shows that the preferred encoding depends on the temporal horizon, update rule, and error metric. At $T=1$, joint encoding gives lower relative $L_2$ and MSE under both updates, whereas separate encoding gives slightly lower relative $L_\infty$. At $T=2$, joint encoding is lower across all three metrics with direct summation, while separate encoding is marginally lower with the GCR-PINN update. At $T=4$, separate encoding improves the global measures in both comparisons, but joint encoding retains the lower relative $L_\infty$ under GCR-PINN. Explicit space--time factorization therefore offers no uniform advantage over the joint LDA representation.

\begin{figure}[H]
\centering
\captionsetup{labelfont={color=black},textfont={color=black}}
\includegraphics[width=.78\textwidth]{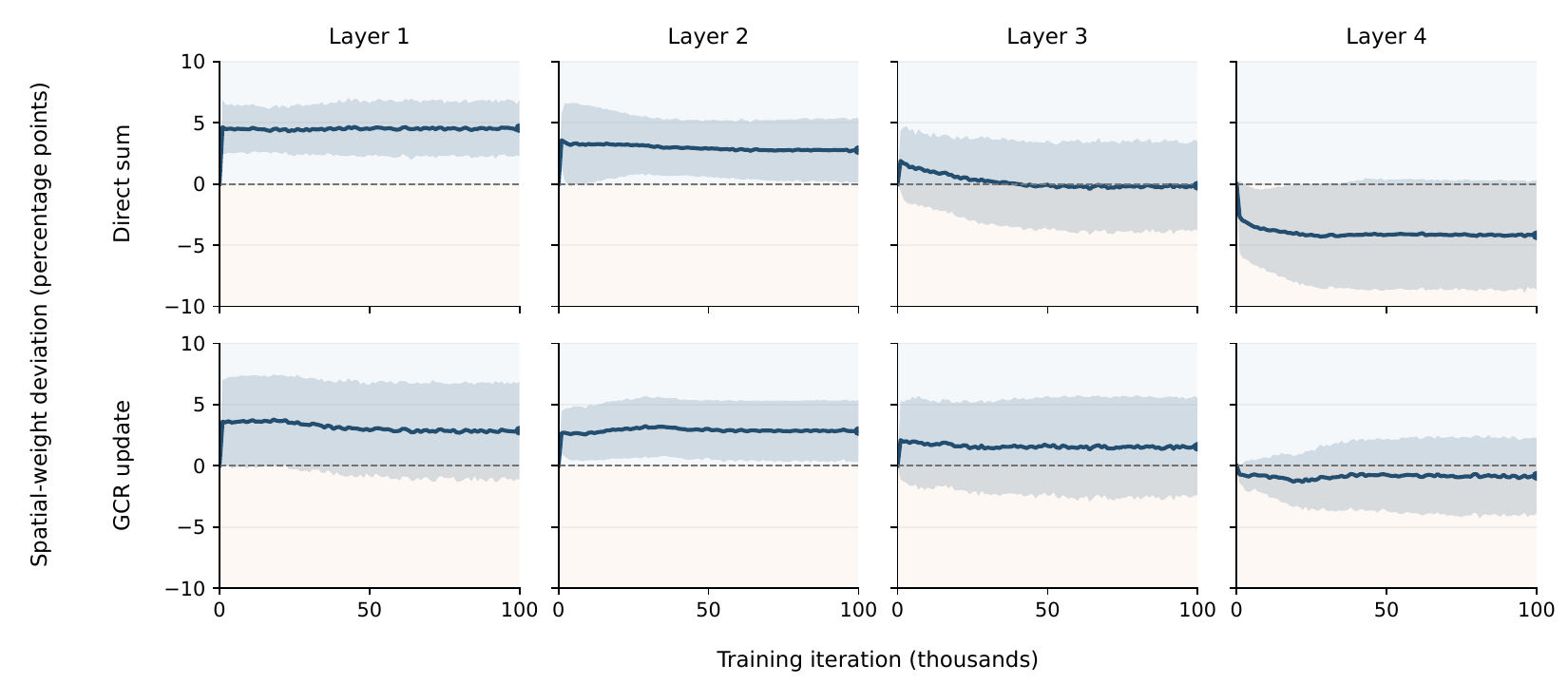}
\caption{Gate allocation in the separate space--time encoder for Klein--Gordon at $T=1$. Columns correspond to hidden layers and rows to direct-sum and GCR-PINN updates. Positive and negative values of $100(\bar{\alpha}_{x}-0.5)$ indicate greater allocation to the spatial and temporal branches, respectively; curves are five-run means and bands span one sample standard deviation.}
\label{fig:kg_separate_gate_trajectory}
\end{figure}

Figure~\ref{fig:kg_separate_gate_trajectory} provides a complementary view of the learned factorization. Under both update rules, the first two layers show a modest preference for the spatial branch, whereas the final layer shifts toward the temporal branch; the intermediate layer remains closer to balanced allocation. The run-to-run bands indicate that these deviations are not uniformly separated from equal weighting, but their depth-dependent pattern confirms that the factorized model does not assign a fixed role to either branch. Combined with Table~\ref{tab:kg_joint_separate}, this result shows that explicit factorization can alter how spatial and temporal information is distributed across depth without producing a consistent accuracy gain. Overall, the crossed-seed and extended-time results preserve the leading mean accuracy of ACR-PINN within the tested variations, while the factorization control supports the joint coordinate encoder as an effective default for the present problems.
}
\FloatBarrier

\subsection{\textcolor{black}{Computational Cost and Complexity}}
\label{sec:computational_cost}
{\color{black}
Because LDA increases the cost of each network update and GCR-PINN evaluates the task gradients separately, iteration-matched results alone do not characterize computational efficiency. We therefore compare the five configurations under a common wall-clock budget on the oscillatory Helmholtz $(4,4)$ problem. The wall-clock budget was fixed a priori at $2462$~s, based on the median projected time for $100{,}000$ updates across the four core methods in synchronized pilot measurements. Table~\ref{tab:h44_time_matched_cost} reports the mean number of completed updates and the resulting prediction accuracy. It also records how many of the five runs reach the predefined criterion of relative $L_2<5\times10^{-2}$; the value in parentheses is the median attainment time among the successful runs. The timing procedure and memory profiles are detailed in~\ref{app:controlled_cost}.

\begin{table}[H]
\centering\color{black}\small\captionsetup{font={color=black}}
\caption{\textcolor{black}{Equal-time accuracy on Helmholtz $(4,4)$ under a wall-clock budget of $2462$~s. Errors are means $\pm$ sample standard deviations over five paired runs; the smallest mean is shown in bold. PM-MLP denotes the parameter-matched MLP.}}
\label{tab:h44_time_matched_cost}
\begin{tabular*}{\textwidth}{@{\extracolsep{\fill}}lccc@{}}
\toprule
Method & Mean updates & Relative $L_2$ & Threshold runs (median s)\\
\midrule
Std-PINN & 284,781 & $(5.68\pm1.21)\times10^{-2}$ & 2/5 (2124)\\
GCR-PINN & 233,246 & $\mathbf{(8.98\pm2.62)\times10^{-3}}$ & 5/5 (116)\\
PM-MLP & 218,161 & $(4.82\pm0.17)\times10^{-2}$ & 4/5 (2282)\\
LDA-PINN & 74,511 & $(6.90\pm0.80)\times10^{-2}$ & 0/5 (--)\\
ACR-PINN & 62,901 & $(1.46\pm0.46)\times10^{-2}$ & 5/5 (294)\\
\bottomrule
\end{tabular*}
\end{table}

The equal-time comparison separates predictive improvement from update throughput. GCR-PINN attains the lowest mean relative $L_2$ error and reaches the prescribed threshold in all five runs, with a median attainment time of $116$~s. ACR-PINN also reaches the threshold in every run, with a median time of $294$~s, despite completing only $62{,}901$ updates on average. Although its mean error is higher than that of GCR-PINN, it remains below those of Std-PINN, PM-MLP, and LDA-PINN, which complete substantially more updates within the same budget. Thus, within this equal-time comparison, GCR-PINN provides the most favorable time--accuracy trade-off, whereas ACR-PINN trades update throughput for the additional representation capacity introduced by LDA. The ACR-PINN accuracy observed here therefore cannot be explained by a larger number of parameter updates.

The two components contribute different sources of overhead. LDA introduces two layer-specific coordinate encoders and a feature-wise gate at each hidden layer, increasing the forward- and backward-pass computation as well as activation storage. GCR-PINN, by contrast, incurs the cost of computing and retaining the gradients of the individual physical tasks. For $P$ trainable parameters and $K$ task losses, storing these gradients requires $O(KP)$ memory, while exhaustive pairwise projection has a worst-case cost of $O(K^2P)$. The empirical profiles in~\ref{app:controlled_cost} reflect both effects: layer-wise encoding accounts for most of the per-update and memory increase, with conflict resolution adding further overhead within either backbone. Under the small number of physical tasks considered here, all configurations remain within the memory capacity of the evaluation hardware. For multi-field or multi-physics systems with larger $K$, task grouping or selective projection would need to be considered to control the quadratic projection cost.}

\FloatBarrier

\section{Conclusion}\label{sec:Conclusion}

{\color{black}
This study presented ACR-PINN as a coordinated architecture--optimization framework for addressing representation limitations and competing physical constraints in PINN training. LDA constructs layer-specific coordinate representations through feature-wise gating and residual fusion, while GCR-PINN treats the PDE, initial-condition, and boundary-condition losses as distinct tasks and conditionally resolves negatively aligned gradients before aggregation. Their integration jointly modifies the learned representation and the parameter update without altering the underlying physical loss formulation.

Across seven benchmark problems evaluated using five matched random seeds, ACR-PINN attains the lowest observed mean errors among the four core architecture--optimizer combinations, with gains that vary across equations and error measures. Comparisons with representative PINN methods further place its accuracy at or near the leading level on the examined problems. The architectural controls show that the LDA improvements cannot be explained by parameter count alone, while interventions on the learned gates demonstrate a measurable dependence of the predictions on the coordinate-fusion pathway. The architecture--optimizer interaction estimates consistently favor the combined formulation, and the task-gradient diagnostics on Burgers and Helmholtz $(4,4)$ associate LDA with lower late-stage conflict rates and smaller conflict-resolution corrections. Collectively, these results indicate that adaptive coordinate representation and conflict-aware gradient updates make complementary contributions to the predictive accuracy of ACR-PINN.

The crossed-seed and extended-time assessments preserve the leading mean accuracy of ACR-PINN within the evaluated variations in sampling, initialization, and temporal horizon. The encoding-factorization study further shows that explicitly separating spatial and temporal representations does not provide a consistent accuracy advantage, supporting joint coordinate encoding as a unified formulation for the problems considered. The equal-time comparison nevertheless reveals an accuracy--cost trade-off: GCR-PINN provides the most favorable time--accuracy balance on Helmholtz $(4,4)$, whereas ACR-PINN retains strong accuracy despite the higher per-update cost introduced by LDA. The present evidence is limited to canonical, predominantly low-dimensional PDEs with a small number of physical tasks, and errors increase as the Klein--Gordon time horizon is extended. Future work should therefore examine more efficient selective conflict-resolution strategies, strongly coupled multi-physics systems, and scalable extensions to high-dimensional and large-scale problems.
}

\section*{Funding information}
The authors gratefully acknowledge the support from the National Natural Science Foundation of China (No.~12371434), the National Key R\&D Program of China (No.~2022YFE03040002), the Foundation of the Sichuan Provincial Department of Science and Technology (No.~2026NSFSC0138), the Key Laboratory of Numerical Simulation of Sichuan Provincial Universities (Nos.~KLNS-2023SZFZ002 and KLNS-2023SZFZ005), and the Key Laboratory of Mathematical Meteorology (No.~2025Z0340).

\section*{Declaration of Competing Interest}
The authors declare that they have no known competing financial interests or personal relationships that could have appeared to influence the work reported in this paper.

\section*{Data Availability}
The data supporting the findings of this study are available from the corresponding author upon reasonable request.

\section*{Code Availability}
The source code is publicly available in the \href{https://github.com/PanChengN/ACR-PINN}{ACR-PINN repository}.

\FloatBarrier
\appendix
\counterwithin{table}{section}
\counterwithin{figure}{section}
\renewcommand{\thetable}{\Alph{section}.\arabic{table}}
\renewcommand{\thefigure}{\Alph{section}.\arabic{figure}}

\section{\textcolor{black}{Benchmark Problems and Evaluation Protocol}}
\label{app:benchmark_definitions}
{\color{black}
\subsection{Governing Equations and Physical Constraints}
For completeness and reproducibility, this subsection collects the governing equations, computational domains, physical constraints, and evaluated fields for the seven benchmarks used in Section~\ref{sec:Experiments}.

\paragraph{Burgers equation}
The viscous Burgers benchmark is defined by
\begin{align*}
\text{PDE:}\quad &u_t+uu_x-\frac{0.01}{\pi}u_{xx}=0,
&& (x,t)\in[-1,1]\times[0,1],\\
\text{initial condition:}\quad &u(x,0)=-\sin(\pi x),\\
\text{boundary conditions:}\quad &u(-1,t)=0,\qquad u(1,t)=0.
\end{align*}

\paragraph{Helmholtz problems}
The two Helmholtz benchmarks are generated from the same manufactured-solution family:
\begin{align*}
\text{PDE:}\quad &\Delta u+u=q,
&& (x,y)\in[-1,1]^2,\\
\text{reference solution:}\quad
&u_\star(x,y)=\sin(a_1\pi x)\sin(a_2\pi y),\\
\text{forcing term:}\quad
&q=\bigl[1-(a_1\pi)^2-(a_2\pi)^2\bigr]u_\star,\\
\text{boundary condition:}\quad &u\big|_{\partial\Omega}=0,\\
\text{parameters:}\quad &(a_1,a_2)\in\{(1,4),(4,4)\}.
\end{align*}

\paragraph{Klein--Gordon equation}
The nonlinear Klein--Gordon benchmark is specified by
\begin{align*}
\text{PDE:}\quad &u_{tt}-u_{xx}+u^3=f,
&& (x,t)\in[0,1]^2,\\
\text{reference solution:}\quad
&u_\star(x,t)=x\cos(5\pi t)+(xt)^3,\\
\text{forcing term:}\quad
&f=-25\pi^2x\cos(5\pi t)+6x^3t-6xt^3+u_\star^3,\\
\text{initial displacement:}\quad &u(x,0)=x,\\
\text{initial velocity:}\quad &u_t(x,0)=0,\\
\text{left boundary:}\quad &u(0,t)=0,\\
\text{right boundary:}\quad &u(1,t)=\cos(5\pi t)+t^3.
\end{align*}
\paragraph{Lid-driven cavity flow}
At $\mathrm{Re}=100$, the steady incompressible flow is governed by
\begin{align*}
\text{$x$-momentum:}\quad
&uu_x+vu_y+p_x-\mathrm{Re}^{-1}\Delta u=0,\\
\text{$y$-momentum:}\quad
&uv_x+vv_y+p_y-\mathrm{Re}^{-1}\Delta v=0,\\
\text{incompressibility:}\quad &u_x+v_y=0,
&& (x,y)\in[0,1]^2,\\
\text{moving lid:}\quad &(u,v)=(1,0), && y=1,\\
\text{stationary walls:}\quad &(u,v)=(0,0),
&& y=0\ \text{or}\ x\in\{0,1\},\\
\text{network representation:}\quad &u=\psi_y,\qquad v=-\psi_x.
\end{align*}
The network outputs $(\psi,p)$, from which the velocity is recovered through the streamfunction. Since the momentum equations depend only on pressure gradients, the additive pressure constant does not affect the solution. The reported error is evaluated for the speed $\sqrt{u^2+v^2}$; pressure is not included in the error metric.

\paragraph{Nonlinear Schr\"odinger equation}
Writing the complex-valued field as $h=u+iv$, the benchmark is defined by
\begin{align*}
\text{PDE:}\quad &ih_t+\tfrac12h_{xx}+|h|^2h=0,
&& (x,t)\in[-5,5]\times[0,\pi/2],\\
\text{initial condition:}\quad &h(x,0)=2\,\mathrm{sech}(x),\\
\text{periodic value:}\quad &h(-5,t)=h(5,t),\\
\text{periodic derivative:}\quad &h_x(-5,t)=h_x(5,t).
\end{align*}
The network outputs the real and imaginary components of $h$ and is evaluated against the numerical reference data (\texttt{NLS.mat}) using $|h_\theta-h|$. Amplitude plots of $|h|$ are identified separately.

\paragraph{Five-dimensional Poisson equation}
The five-dimensional scalar problem is
\begin{align*}
\text{PDE:}\quad &-\Delta u=f,
&& \boldsymbol{x}\in[0,1]^5,\\
\text{reference solution:}\quad
&u_\star(\boldsymbol{x})=\sum_{k=1}^{5}\sin(\pi x_k/2),\\
\text{forcing term:}\quad &f=\frac{\pi^2}{4}u_\star,\\
\text{boundary condition:}\quad &u\big|_{\partial\Omega}=u_\star.
\end{align*}
All metrics use 100,000 independently generated evaluation points.
}

\FloatBarrier

\subsection{\textcolor{black}{Training and Evaluation Settings}}
\label{app:network_evaluation}
{\color{black}
The core comparison uses the problem-dependent collocation and iteration budgets summarized in Table~\ref{tab:core_protocol}. Because the training points are regenerated at every iteration, the reported sample counts are per-update quantities rather than totals accumulated over training.
\begin{table}[htbp]
\centering\color{black}
\small
\captionsetup{font={color=black}}
\caption{Per-update sampling and total iteration budgets for the seven benchmark cases. The two numbers in the cavity boundary entry correspond to moving-lid and stationary-wall points, respectively.}
\label{tab:core_protocol}
\begin{tabular*}{\textwidth}{@{\extracolsep{\fill}}lcccc@{}}
\toprule
Benchmark case & Interior points & Boundary points & Initial points & Iterations \\
\midrule
Burgers & 20,000 & 500 & 500 & 40,000 \\
Helmholtz $(1,4)$ & 10,000 & 400 & -- & 100,000 \\
Helmholtz $(4,4)$ & 20,000 & 400 & -- & 100,000 \\
Klein--Gordon & 10,000 & 200 & 200 & 100,000 \\
Lid-driven cavity & 1,000 & $300+300$ & -- & 100,000 \\
Nonlinear Schr\"odinger & 20,000 & 500 & 500 & 100,000 \\
5D Poisson & 20,000 & 5,000 & -- & 100,000 \\
\bottomrule
\end{tabular*}
\end{table}

Randomized Latin-hypercube sampling (LHS) regenerates the collocation points at every iteration for all four core methods. Optimization uses Adam with a 100-iteration linear warm-up to a learning rate of $10^{-3}$, followed by cosine decay to $10^{-4}$. Each backbone contains four hidden layers of width 50 with $\tanh$ activation; LDA additionally introduces its layer-specific coordinate encoders and feature-wise gates. The problem-dependent input--output dimensions, evaluation sets, and reported fields are summarized in Table~\ref{tab:network_evaluation}. The two outputs of the nonlinear Schr\"odinger model represent the real and imaginary components, whereas those of the cavity model represent the streamfunction and pressure.
\begin{table}[htbp]
\centering\color{black}\small\captionsetup{font={color=black}}
\caption{Problem-dependent input--output dimensions, evaluation sets, and reported fields. Evaluation points are independent of the training samples.}
\label{tab:network_evaluation}
\setlength{\tabcolsep}{4pt}
\begin{tabular}{lcccc}
\toprule
Benchmark case & Input dim. & Output dim. & Evaluation set & Evaluated field\\
\midrule
Burgers & 2 & 1 & $256\times100$ & $u$\\
Helmholtz $(1,4)$ & 2 & 1 & $200\times200$ & $u$\\
Helmholtz $(4,4)$ & 2 & 1 & $200\times200$ & $u$\\
Klein--Gordon & 2 & 1 & $200\times200$ & $u$\\
Lid-driven cavity & 2 & 2 & $100\times100$ & $\sqrt{u^2+v^2}$\\
Nonlinear Schr\"odinger & 2 & 2 & $256\times201$ & $h=u+iv$\\
5D Poisson & 5 & 1 & 100,000 LHS points & $u$\\
\bottomrule
\end{tabular}
\end{table}
\FloatBarrier

All evaluation points are excluded from training and model selection. Errors for problems with structured reference grids are computed on the corresponding discrete samples; the five-dimensional Poisson problem is instead evaluated on an independent Latin-hypercube sample.
}

\FloatBarrier

\section{\textcolor{black}{Task-Resolved Gradient Diagnostics}}
\label{app:gradient_diagnostics}
{\color{black}
Figures~\ref{fig:task_gradient_norms}--\ref{fig:helmholtz_task_pair} decompose the aggregate gradient diagnostics in Section~\ref{sec:gradient_geometry} by physical task and task pair, thereby identifying which interactions among the physical constraints account for the observed changes over training.

\begin{figure}[htbp]
\centering
\color{black}
\captionsetup{font={color=black}}
\includegraphics[width=.90\textwidth]{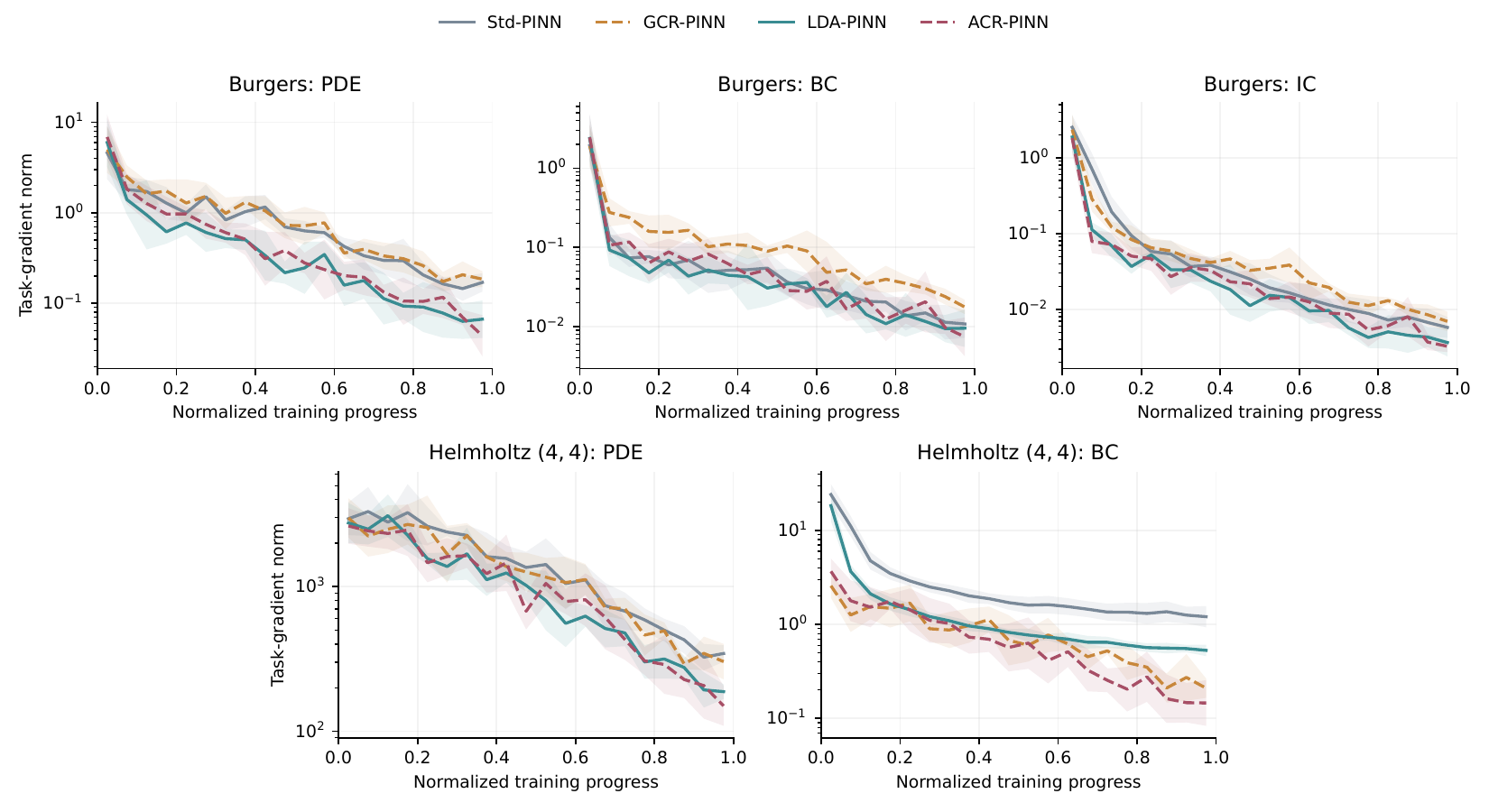}
\caption{Pre-projection task-gradient norms for Burgers and Helmholtz $(4,4)$. The upper row resolves the PDE, boundary-condition, and initial-condition tasks for Burgers, while the lower row resolves the PDE and boundary-condition tasks for stationary Helmholtz. Curves denote geometric means over five runs, and bands span one sample standard deviation in $\log_{10}$ space.}
\label{fig:task_gradient_norms}
\end{figure}

\begin{figure}[htbp]
\centering
\color{black}
\captionsetup{font={color=black}}
\includegraphics[width=.78\textwidth]{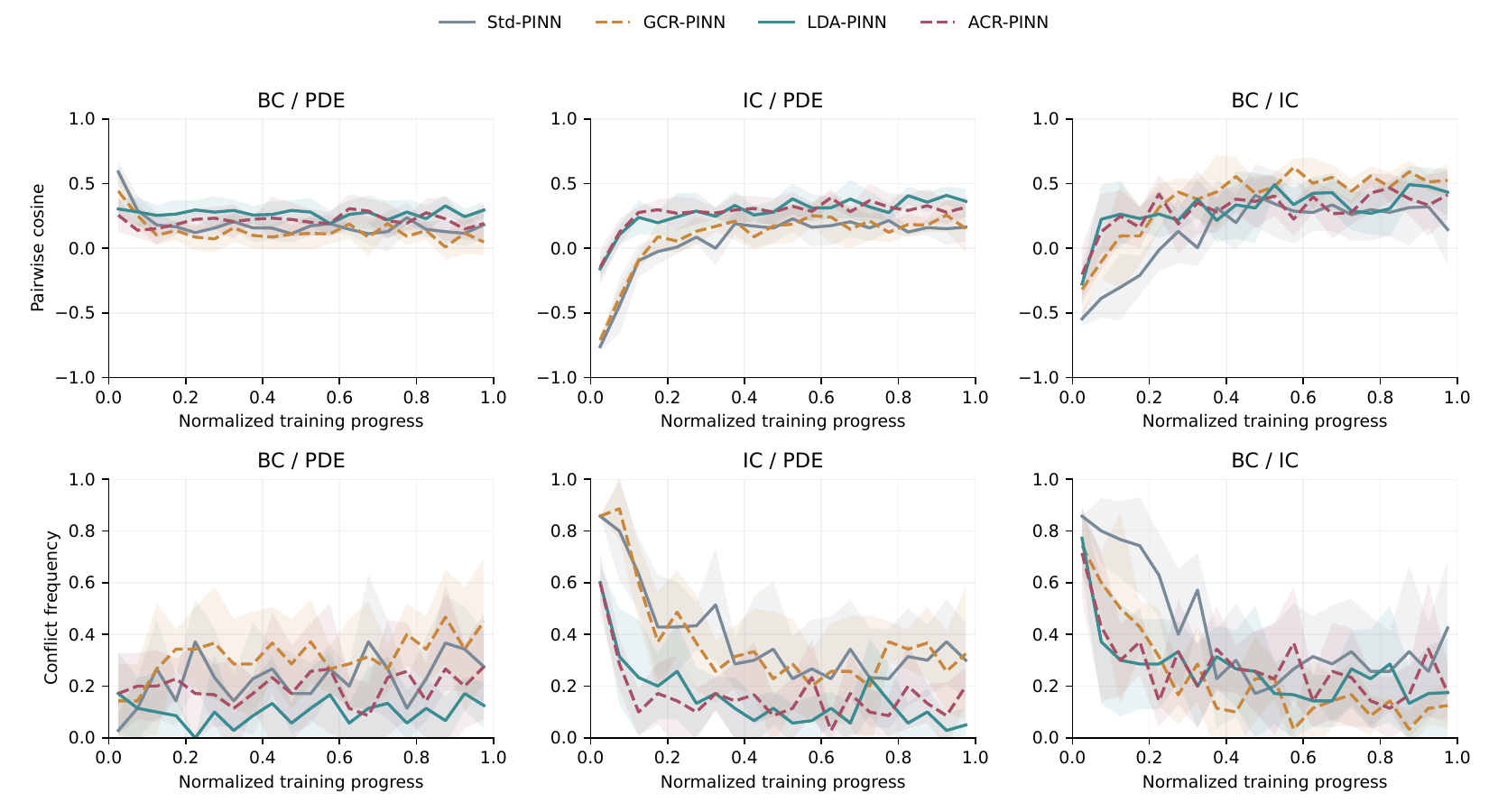}
\caption{Pairwise task-gradient geometry for Burgers. Columns correspond to the boundary--PDE, initial--PDE, and boundary--initial pairs; the upper and lower rows show cosine similarity and conflict frequency, respectively. Curves denote five-run means with pointwise 95\% bootstrap intervals.}
\label{fig:burgers_task_pairs}
\end{figure}

\begin{figure}[htbp]
\centering
\color{black}
\captionsetup{font={color=black}}
\includegraphics[width=.44\textwidth]{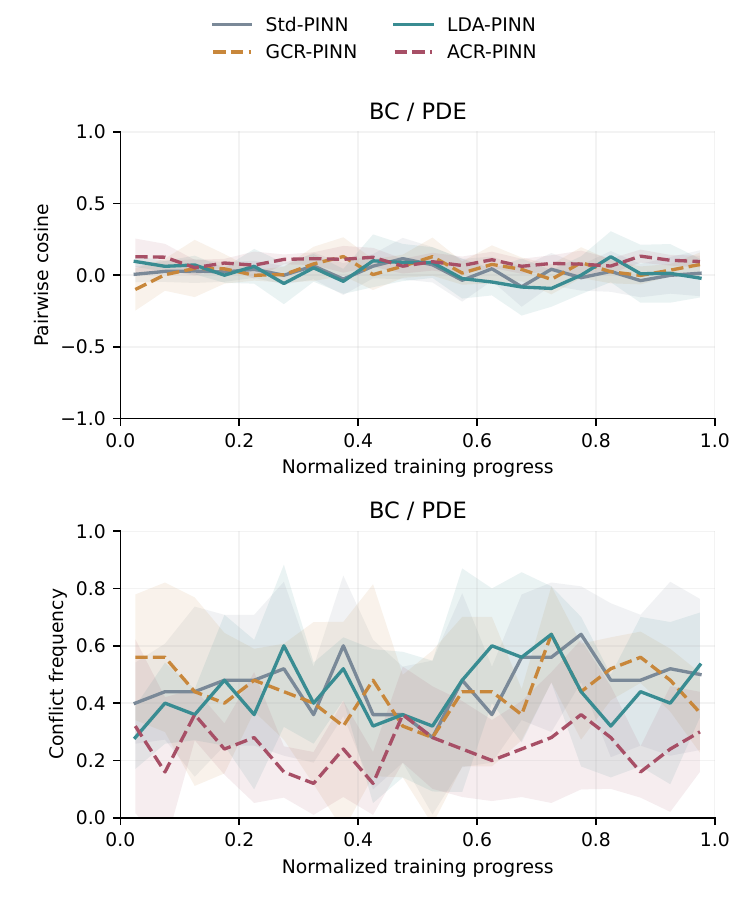}
\caption{PDE--boundary task-gradient geometry for Helmholtz $(4,4)$. The upper and lower panels show cosine similarity and conflict frequency, respectively. Curves denote five-run means with pointwise 95\% bootstrap intervals.}
\label{fig:helmholtz_task_pair}
\end{figure}
\FloatBarrier
Figure~\ref{fig:task_gradient_norms} shows that the task-gradient norms decrease under all four methods, but the relative trajectories vary by physical constraint. For Burgers, the LDA-based models generally reach lower late-stage magnitudes for the PDE and initial-condition tasks. The pairwise diagnostics in Fig.~\ref{fig:burgers_task_pairs} provide the more direct measure of gradient compatibility: LDA-PINN and ACR-PINN tend to maintain higher cosine similarities and lower conflict frequencies for the boundary--PDE and initial--PDE pairs, whereas the differences for the boundary--initial pair are less uniform. The aggregate improvement on Burgers is therefore driven primarily by changes in the directional relationships involving the PDE task.

For Helmholtz $(4,4)$, the PDE-gradient norms follow broadly similar decreasing trends, while the boundary-gradient trajectories separate more clearly: Std-PINN remains highest, LDA-PINN occupies an intermediate range, and the two conflict-resolved methods decrease further, with ACR-PINN reaching the lowest late-stage level. Gradient magnitude alone, however, does not determine compatibility. Figure~\ref{fig:helmholtz_task_pair} shows that the PDE and boundary gradients fluctuate around near-orthogonal alignment for all four methods, while their conflict frequencies differ substantially. ACR-PINN maintains a lower conflict frequency through most of training than the other configurations, indicating that the combined formulation primarily reduces the occurrence of negatively aligned updates rather than inducing a large displacement of the mean cosine.

To quantify these temporal patterns, the early and late stages are defined as the first and final thirds of normalized training progress. Each diagnostic is first averaged within stage and run, after which the matched LDA-minus-MLP contrasts are computed in Table~\ref{tab:gradient_stage_contrasts}.
\begin{table}[H]
\centering\color{black}\scriptsize\captionsetup{font={color=black}}
\caption{LDA-minus-MLP contrasts in task-gradient geometry during the early and late stages of training. Entries are mean differences over five matched runs with marginal 95\% bootstrap intervals. Here $C$, $\bar c$, $S$, and $P$ denote conflict rate, mean pairwise cosine, gradient-norm spread, and projection ratio, respectively; $P$ is defined only under the GCR-PINN update.}
\label{tab:gradient_stage_contrasts}
\resizebox{\textwidth}{!}{\begin{tabular}{lllcccc}
\toprule
PDE & Phase & Update rule & $\Delta C$ & $\Delta\bar c$ & $\Delta S$ & $\Delta P$ \\
\midrule
Burgers & Early & Direct sum & $-0.240\;[-0.268,-0.216]$ & $+0.253\;[+0.235,+0.274]$ & $+0.011\;[-0.060,+0.096]$ & -- \\
& Early & GCR-PINN & $-0.160\;[-0.200,-0.120]$ & $+0.128\;[+0.087,+0.172]$ & $+0.075\;[+0.040,+0.112]$ & $-0.193\;[-0.212,-0.174]$ \\
& Late & Direct sum & $-0.157\;[-0.232,-0.081]$ & $+0.137\;[+0.064,+0.203]$ & $-0.111\;[-0.180,-0.042]$ & -- \\
& Late & GCR-PINN & $-0.078\;[-0.128,-0.026]$ & $+0.036\;[-0.008,+0.082]$ & $-0.110\;[-0.202,+0.007]$ & $-0.062\;[-0.093,-0.022]$ \\
\midrule
Helmholtz $(4,4)$ & Early & Direct sum & $-0.029\;[-0.153,+0.059]$ & $+0.011\;[-0.029,+0.063]$ & $+0.186\;[+0.048,+0.310]$ & -- \\
& Early & GCR-PINN & $-0.224\;[-0.300,-0.141]$ & $+0.087\;[+0.041,+0.134]$ & $-0.116\;[-0.218,-0.032]$ & $-0.076\;[-0.105,-0.048]$ \\
& Late & Direct sum & $-0.053\;[-0.235,+0.159]$ & $+0.001\;[-0.131,+0.133]$ & $+0.130\;[+0.091,+0.187]$ & -- \\
& Late & GCR-PINN & $-0.218\;[-0.306,-0.124]$ & $+0.060\;[+0.024,+0.096]$ & $-0.016\;[-0.192,+0.116]$ & $-0.045\;[-0.055,-0.037]$ \\
\bottomrule
\end{tabular}}
\end{table}

The stage-wise contrasts quantify the distinctions visible in the trajectories. For Burgers, LDA generally reduces the conflict rate and improves directional alignment under both direct summation and GCR-PINN. For Helmholtz $(4,4)$, the changes are limited under direct summation but become more pronounced under GCR-PINN, for which both the conflict rate and projection ratio decrease. These comparisons show that LDA does not merely rescale all task gradients uniformly; it alters the directional relationships among specific physical constraints. This effect appears under both update rules for Burgers and is most pronounced when LDA is combined with GCR-PINN for Helmholtz $(4,4)$, providing task-level evidence that adaptive feature representation and conflict-resolved optimization make complementary contributions to gradient compatibility.
}

\FloatBarrier

\section{\textcolor{black}{Configuration of the Representative Baselines}}
\label{app:baseline_configuration}
{\color{black}
For the controlled comparison in Section~\ref{sec:strong_baselines}, all methods use two inputs, one output, and four hidden layers of width 50 with $\tanh$ activation unless their defining architecture requires otherwise. Table~\ref{tab:strong_baseline_configuration} records the trainable parameter counts and method-specific architectural or optimization settings. Dynamic weighting states and the fixed Fourier matrix are excluded from the parameter counts.

\begin{table}[H]
\centering
\color{black}
\captionsetup{font={color=black}}
\caption{Architectural and optimization configurations used in the fixed-sampling baseline comparison. Parameter counts apply to both Burgers and Helmholtz $(4,4)$ and exclude the fixed Fourier matrix, optimizer states, and dynamic weighting states.}
\label{tab:strong_baseline_configuration}
\scriptsize
\setlength{\tabcolsep}{3pt}
\begin{tabular}{@{}>{\raggedright\arraybackslash}p{.18\textwidth}>{\centering\arraybackslash}p{.13\textwidth}>{\raggedright\arraybackslash}p{.63\textwidth}@{}}
\toprule
Method & Parameters & Method-specific configuration \\
\midrule
Std-PINN & 7,851 & Summed physical losses. \\
GCR-PINN & 7,851 & Conflict-resolved gradient update. \\
LDA-PINN & 59,655 & LDA backbone; summed losses. \\
ACR-PINN & 59,655 & LDA backbone; conflict-resolved gradient update. \\
M-MLP+LRA & 8,151 & Shared input encoders with feature-wise mixing~\cite{wang2021understanding}; weights updated every 10 iterations with exponential moving-average coefficient 0.9. \\
FF-PINN & 12,751 & Fixed Gaussian Fourier map $[\sin(2\pi\boldsymbol{x}B),\cos(2\pi\boldsymbol{x}B)]$ with 50 frequencies; scale 1 selected from $\{1,5\}$; summed losses~\cite{tancik2020fourier}. \\
RBA-PINN & 7,851 & Pointwise residual attention~\cite{anagnostopoulos2024rba}; $\eta=0.001$, $\gamma=0.999$; zero initialization. \\
BRDR-PINN & 7,851 & Residual-decay weighting~\cite{chen2025brdr}; $\beta_c=\beta_w=0.999$; unit initialization. \\
\bottomrule
\end{tabular}
\end{table}

}

\FloatBarrier

\section{\textcolor{black}{Computational Resource Profile}}
\label{app:controlled_cost}
{\color{black}
Per-step runtime and peak allocated memory were measured on an NVIDIA GeForce RTX 4090 D using PyTorch 2.0.1+cu117 and CUDA 11.7. Each value summarizes three independent repetitions, each comprising 100 warm-up steps followed by 200 measured steps. Data generation and solution evaluation were excluded to isolate the computational cost of the parameter update.

The common wall-clock budget used in Section~\ref{sec:computational_cost} was derived independently from synchronized, uncontended pilot measurements of the complete Helmholtz $(4,4)$ training loop, including per-iteration resampling. The measured full-loop times were $11.492$, $12.657$, $36.577$, and $46.968$~ms per update for Std-PINN, GCR-PINN, LDA-PINN, and ACR-PINN, respectively. Before the equal-time errors were examined, the median projected duration of $100{,}000$ updates was computed from the two central measurements as $2461.7$~s and rounded to a common budget of $2462$~s.

\begin{table}[H]
\centering\color{black}\scriptsize\captionsetup{font={color=black}}
\caption{Per-step computational cost across the seven benchmark cases. Panel (a) reports the mean runtime $\pm$ one sample standard deviation over three repetitions, in milliseconds, and panel (b) reports peak allocated device memory, in GiB. PM-MLP denotes the parameter-matched MLP.}
\label{tab:controlled_cost_full}
\setlength{\tabcolsep}{4pt}
\renewcommand{\arraystretch}{1.05}
\textit{(a) Step time (ms)}\\[2pt]
\begin{tabular*}{\textwidth}{@{\extracolsep{\fill}}lccccc@{}}
\toprule
Benchmark case & Std-PINN & GCR-PINN & LDA-PINN & ACR-PINN & PM-MLP\\
\midrule
Burgers & $5.13\pm0.84$ & $12.17\pm1.46$ & $28.52\pm0.18$ & $46.90\pm9.73$ & $8.71\pm1.22$\\
Helmholtz $(1,4)$ & $6.30\pm1.83$ & $9.93\pm2.00$ & $31.01\pm3.21$ & $45.79\pm6.36$ & $6.33\pm1.59$\\
Helmholtz $(4,4)$ & $8.07\pm0.58$ & $8.66\pm1.63$ & $32.09\pm0.98$ & $42.82\pm1.57$ & $9.42\pm0.46$\\
Klein--Gordon & $10.63\pm1.22$ & $19.54\pm2.37$ & $37.88\pm2.02$ & $85.70\pm2.00$ & $9.82\pm2.33$\\
Lid-driven cavity & $19.02\pm1.37$ & $28.40\pm3.19$ & $135.75\pm11.58$ & $165.68\pm11.58$ & $18.75\pm1.04$\\
5D Poisson & $12.92\pm1.61$ & $13.96\pm1.99$ & $63.70\pm4.44$ & $74.41\pm4.48$ & $18.61\pm0.94$\\
Nonlinear Schr\"odinger & $12.80\pm2.65$ & $23.14\pm1.79$ & $57.18\pm1.15$ & $87.47\pm6.58$ & $17.63\pm0.47$\\
\bottomrule
\end{tabular*}

\vspace{5pt}
\textit{(b) Peak allocated memory (GiB)}\\[2pt]
\begin{tabular*}{\textwidth}{@{\extracolsep{\fill}}lccccc@{}}
\toprule
Benchmark case & Std-PINN & GCR-PINN & LDA-PINN & ACR-PINN & PM-MLP\\
\midrule
Burgers & 0.21 & 0.30 & 1.35 & 2.04 & 0.53\\
Helmholtz $(1,4)$ & 0.16 & 0.24 & 1.01 & 1.62 & 0.40\\
Helmholtz $(4,4)$ & 0.30 & 0.47 & 2.00 & 3.21 & 0.78\\
Klein--Gordon & 0.16 & 0.24 & 1.02 & 1.62 & 0.40\\
Lid-driven cavity & 0.08 & 0.12 & 0.45 & 0.75 & 0.19\\
5D Poisson & 0.58 & 1.02 & 3.87 & 6.91 & 1.52\\
Nonlinear Schr\"odinger & 0.35 & 0.54 & 2.31 & 3.75 & 0.91\\
\bottomrule
\end{tabular*}
\end{table}

Across the seven benchmarks, layer-wise coordinate encoding accounts for the larger increase in both runtime and memory, while conflict resolution introduces additional overhead within either backbone. Consequently, ACR-PINN has the highest per-step resource demand among the four core methods, consistent with the accuracy--cost trade-off discussed in Section~\ref{sec:computational_cost}.
}

\FloatBarrier
\bibliographystyle{elsarticle-num}
\bibliography{myfile}

@article{raissi2019physics,
  title={Physics-informed neural networks: A deep learning framework for solving forward and inverse problems involving nonlinear partial differential equations},
  author={Raissi, Maziar and Perdikaris, Paris and Karniadakis, George E},
  journal={Journal of Computational Physics},
  volume={378},
  pages={686--707},
  year={2019},
  publisher={Elsevier}
}

@article{karniadakis2021physics,
  title={Physics-informed machine learning},
  author={Karniadakis, George Em and Kevrekidis, Ioannis G and Lu, Lu and Perdikaris, Paris and Wang, Sifan and Yang, Liu},
  journal={Nature Reviews Physics},
  volume={3},
  number={6},
  pages={422--440},
  year={2021},
  publisher={Nature Publishing Group UK London}
}

@article{cai2021physics,
  title={Physics-informed neural networks for heat transfer problems},
  author={Cai, Shengze and Wang, Zhicheng and Wang, Sifan and Perdikaris, Paris and Karniadakis, George Em},
  journal={Journal of Heat Transfer},
  volume={143},
  number={6},
  pages={060801},
  year={2021},
  publisher={American Society of Mechanical Engineers}
}

@article{haghighat2021physics,
  title={A physics-informed deep learning framework for inversion and surrogate modeling in solid mechanics},
  author={Haghighat, Ehsan and Raissi, Maziar and Moure, Adrian and Gomez, Hector and Juanes, Ruben},
  journal={Computer Methods in Applied Mechanics and Engineering},
  volume={379},
  pages={113741},
  year={2021},
  publisher={Elsevier}
}

@article{jin2021nsfnets,
  title={{NSFnets} ({Navier--Stokes} flow nets): Physics-informed neural networks for the incompressible {Navier--Stokes} equations},
  author={Jin, Xiaowei and Cai, Shengze and Li, Hui and Karniadakis, George Em},
  journal={Journal of Computational Physics},
  volume={426},
  pages={109951},
  year={2021},
  publisher={Elsevier}
}

@article{jagtap2022physics,
  title={Physics-informed neural networks for inverse problems in supersonic flows},
  author={Jagtap, Ameya D and Mao, Zhiping and Adams, Nikolaus and Karniadakis, George Em},
  journal={Journal of Computational Physics},
  volume={466},
  pages={111402},
  year={2022},
  publisher={Elsevier}
}

@article{xu2023physics,
  title={Physics-informed neural networks for studying heat transfer in porous media},
  author={Xu, Jiaxuan and Wei, Han and Bao, Hua},
  journal={International Journal of Heat and Mass Transfer},
  volume={217},
  pages={124671},
  year={2023},
  publisher={Elsevier}
}

@article{hu2024physics,
  title={Physics-informed neural networks ({PINN}) for computational solid mechanics: Numerical frameworks and applications},
  author={Hu, Haoteng and Qi, Lehua and Chao, Xujiang},
  journal={Thin-Walled Structures},
  volume={205},
  pages={112495},
  year={2024},
  publisher={Elsevier}
}

@article{baydin2018automatic,
  title={Automatic differentiation in machine learning: a survey},
  author={Baydin, Atilim Gunes and Pearlmutter, Barak A and Radul, Alexey Andreyevich and Siskind, Jeffrey Mark},
  journal={Journal of Machine Learning Research},
  volume={18},
  number={153},
  pages={1--43},
  year={2018}
}

@inproceedings{krishnapriyan2021characterizing,
  title={Characterizing possible failure modes in physics-informed neural networks},
  author={Krishnapriyan, Aditi and Gholami, Amir and Zhe, Shandian and Kirby, Robert and Mahoney, Michael W},
  booktitle={Advances in Neural Information Processing Systems},
  volume={34},
  pages={26548--26560},
  year={2021}
}

@article{jagtap2020adaptive,
  title={Adaptive activation functions accelerate convergence in deep and physics-informed neural networks},
  author={Jagtap, Ameya D and Kawaguchi, Kenji and Karniadakis, George Em},
  journal={Journal of Computational Physics},
  volume={404},
  pages={109136},
  year={2020},
  publisher={Elsevier}
}

@article{wang2022and,
  title={When and why {PINNs} fail to train: A neural tangent kernel perspective},
  author={Wang, Sifan and Yu, Xinling and Perdikaris, Paris},
  journal={Journal of Computational Physics},
  volume={449},
  pages={110768},
  year={2022},
  publisher={Elsevier}
}

@article{song2024loss,
  title={Loss-attentional physics-informed neural networks},
  author={Song, Yanjie and Wang, He and Yang, He and Taccari, Maria Luisa and Chen, Xiaohui},
  journal={Journal of Computational Physics},
  volume={501},
  pages={112781},
  year={2024},
  publisher={Elsevier}
}

@article{maddu2022inverse,
  title={Inverse Dirichlet weighting enables reliable training of physics informed neural networks},
  author={Maddu, Suryanarayana and Sturm, Dominik and M{\"u}ller, Christian L and Sbalzarini, Ivo F},
  journal={Machine Learning: Science and Technology},
  volume={3},
  number={1},
  pages={015026},
  year={2022},
  publisher={IOP Publishing}
}

@article{deguchi2023dynamic,
  title={Dynamic \& norm-based weights to normalize imbalance in back-propagated gradients of physics-informed neural networks},
  author={Deguchi, Shota and Asai, Mitsuteru},
  journal={Journal of Physics Communications},
  volume={7},
  number={7},
  pages={075005},
  year={2023},
  publisher={IOP Publishing}
}

@article{yu2022gradient,
  title={Gradient-enhanced physics-informed neural networks for forward and inverse {PDE} problems},
  author={Yu, Jeremy and Lu, Lu and Meng, Xuhui and Karniadakis, George Em},
  journal={Computer Methods in Applied Mechanics and Engineering},
  volume={393},
  pages={114823},
  year={2022},
  publisher={Elsevier}
}

@article{liu2021dual,
  title={A dual-dimer method for training physics-constrained neural networks with minimax architecture},
  author={Liu, Dehao and Wang, Yan},
  journal={Neural Networks},
  volume={136},
  pages={112--125},
  year={2021},
  publisher={Elsevier}
}

@article{wang2021understanding,
  title={Understanding and mitigating gradient flow pathologies in physics-informed neural networks},
  author={Wang, Sifan and Teng, Yujun and Perdikaris, Paris},
  journal={SIAM Journal on Scientific Computing},
  volume={43},
  number={5},
  pages={A3055--A3081},
  year={2021},
  publisher={SIAM}
}

@article{wang2024practical,
  title={A practical {PINN} framework for multi-scale problems with multi-magnitude loss terms},
  author={Wang, Yong and Yao, Yanzhong and Guo, Jiawei and Gao, Zhiming},
  journal={Journal of Computational Physics},
  volume={510},
  pages={113112},
  year={2024},
  publisher={Elsevier}
}

@article{niu2025improved,
  title={Improved physics-informed neural network in mitigating gradient-related failures},
  author={Niu, Pancheng and Guo, Jun and Chen, Yongming and Zhou, Yuqian and Feng, Minfu and Shi, Yanchao},
  journal={Neurocomputing},
  volume={638},
  pages={130167},
  year={2025},
  publisher={Elsevier}
}

@inproceedings{yu2020gradient,
  title={Gradient surgery for multi-task learning},
  author={Yu, Tianhe and Kumar, Saurabh and Gupta, Abhishek and Levine, Sergey and Hausman, Karol and Finn, Chelsea},
  booktitle={Advances in Neural Information Processing Systems},
  volume={33},
  pages={5824--5836},
  year={2020}
}

@inproceedings{liu2021conflict,
  title={Conflict-averse gradient descent for multi-task learning},
  author={Liu, Bo and Liu, Xingchao and Jin, Xiaojie and Stone, Peter and Liu, Qiang},
  booktitle={Advances in Neural Information Processing Systems},
  volume={34},
  pages={18878--18890},
  year={2021}
}

@inproceedings{liu2024kan,
  title={{KAN}: {Kolmogorov--Arnold} Networks},
  author={Liu, Ziming and Wang, Yixuan and Vaidya, Sachin and Ruehle, Fabian and Halverson, James and Solja{\v{c}}i{\'c}, Marin and Hou, Thomas Y. and Tegmark, Max},
  booktitle={International Conference on Learning Representations},
  year={2025},
  eprint={2404.19756},
  archivePrefix={arXiv},
  primaryClass={cs.LG},
  url={https://arxiv.org/abs/2404.19756}
}

@inproceedings{tancik2020fourier,
  title={{Fourier} features let networks learn high frequency functions in low dimensional domains},
  author={Tancik, Matthew and Srinivasan, Pratul and Mildenhall, Ben and Fridovich-Keil, Sara and Raghavan, Nithin and Singhal, Utkarsh and Ramamoorthi, Ravi and Barron, Jonathan and Ng, Ren},
  booktitle={Advances in Neural Information Processing Systems},
  volume={33},
  pages={7537--7547},
  year={2020}
}

@article{rasht2022physics,
  title={Physics-informed neural networks ({PINNs}) for wave propagation and full waveform inversions},
  author={Rasht-Behesht, Majid and Huber, Christian and Shukla, Khemraj and Karniadakis, George Em},
  journal={Journal of Geophysical Research: Solid Earth},
  volume={127},
  number={5},
  pages={e2021JB023120},
  year={2022},
  publisher={Wiley Online Library}
}

@article{alkhadhr2023wave,
  title={Wave equation modeling via physics-informed neural networks: Models of soft and hard constraints for initial and boundary conditions},
  author={Alkhadhr, Shaikhah and Almekkawy, Mohamed},
  journal={Sensors},
  volume={23},
  number={5},
  pages={2792},
  year={2023},
  publisher={MDPI}
}

@article{chen2024wavenets,
  title={{WaveNets}: physics-informed neural networks for full-field recovery of rotational flow beneath large-amplitude periodic water waves},
  author={Chen, Lin and Li, Ben and Luo, Chenyi and Lei, Xiaoming},
  journal={Engineering with Computers},
  volume={40},
  number={5},
  pages={2819--2839},
  year={2024},
  publisher={Springer}
}

@article{urban2025unveiling,
  title={Unveiling the optimization process of physics informed neural networks: How accurate and competitive can PINNs be?},
  author={Urb{\'a}n, Jorge F and Stefanou, Petros and Pons, Jos{\'e} A},
  journal={Journal of Computational Physics},
  volume={523},
  pages={113656},
  year={2025},
  publisher={Elsevier}
}

@article{luo2025physics,
  title={Physics-informed neural networks for {PDE} problems: a comprehensive review},
  author={Luo, Kuang and Zhao, Jingshang and Wang, Yingping and Li, Jiayao and Wen, Junjie and Liang, Jiong and Soekmadji, Henry and Liao, Shaolin},
  journal={Artificial Intelligence Review},
  volume={58},
  number={10},
  pages={323},
  year={2025},
  publisher={Springer}
}

@inproceedings{vaswani2017attention,
  title={Attention is all you need},
  author={Vaswani, Ashish and Shazeer, Noam and Parmar, Niki and Uszkoreit, Jakob and Jones, Llion and Gomez, Aidan N and Kaiser, {\L}ukasz and Polosukhin, Illia},
  booktitle={Advances in Neural Information Processing Systems},
  volume={30},
  pages={5998--6008},
  year={2017}
}

@inproceedings{sener2018multitask,
  title={Multi-task learning as multi-objective optimization},
  author={Sener, Ozan and Koltun, Vladlen},
  booktitle={Advances in Neural Information Processing Systems},
  volume={31},
  pages={527--538},
  year={2018}
}

@inproceedings{chen2018gradnorm,
  title={{GradNorm}: Gradient normalization for adaptive loss balancing in deep multitask networks},
  author={Chen, Zhao and Badrinarayanan, Vijay and Lee, Chen-Yu and Rabinovich, Andrew},
  booktitle={Proceedings of the 35th International Conference on Machine Learning},
  series={Proceedings of Machine Learning Research},
  volume={80},
  pages={794--803},
  year={2018}
}

@article{anagnostopoulos2024rba,
  title={Residual-based attention in physics-informed neural networks},
  author={Anagnostopoulos, Sokratis J. and Toscano, Juan Diego and Stergiopulos, Nikolaos and Karniadakis, George Em},
  journal={Computer Methods in Applied Mechanics and Engineering},
  volume={421},
  pages={116805},
  year={2024},
  doi={10.1016/j.cma.2024.116805}
}

@article{chen2025brdr,
  title={Self-adaptive weights based on balanced residual decay rate for physics-informed neural networks and deep operator networks},
  author={Chen, Wenqian and Howard, Amanda A. and Stinis, Panos},
  journal={Journal of Computational Physics},
  volume={542},
  pages={114226},
  year={2025},
  doi={10.1016/j.jcp.2025.114226}
}
\end{document}